\documentclass{article}

\usepackage{arxiv}

\usepackage{amsmath}
\usepackage{array}

\usepackage[utf8]{inputenc} 
\usepackage[T1]{fontenc}    
\usepackage[hidelinks]{hyperref}       
\usepackage{url}            
\usepackage{booktabs}       
\usepackage{amsfonts}       
\usepackage{nicefrac}       
\usepackage{microtype}      
\usepackage{cleveref}       
\usepackage{lipsum}         
\usepackage{graphicx}
\usepackage[numbers]{natbib}
\usepackage{doi}

\usepackage{subcaption}
\usepackage{algorithm, algpseudocode}
\usepackage{threeparttable}
\usepackage{multicol}

\newcolumntype{L}[1]{>{\raggedright\let\newline\\\arraybackslash\hspace{0pt}}m{#1}}
\newcolumntype{C}[1]{>{\centering\let\newline\\\arraybackslash\hspace{0pt}}m{#1}}
\newcolumntype{R}[1]{>{\raggedleft\let\newline\\\arraybackslash\hspace{0pt}}m{#1}}

\title{Neighbor displacement-based enhanced synthetic oversampling for multiclass imbalanced data}

\date{}

\newif\ifuniqueAffiliation
\uniqueAffiliationtrue

\usepackage{authblk}

\newbox{\orcid}\sbox{\orcid}{\includegraphics[scale=0.06]{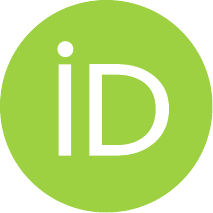}} 
\author[1,2]{%
	\href{https://orcid.org/0000-0002-8013-7423}{\usebox{\orcid}\hspace{1mm}I Made Putrama\thanks{\texttt{putrama.imade@edu.bme.hu}}}%
}
\author[1]{%
	Peter Martinek\thanks{\texttt{martinek.peter@vik.bme.hu}}%
}
\affil[1]{\scriptsize{Department of Electronics Technology, Faculty of Electrical Engineering and Informatics, Budapest University of Technology and Economics, Budapest, Hungary}}
\affil[2]{\scriptsize{Department of Informatics, Faculty of Engineering and Vocational, Universitas Pendidikan Ganesha, Singaraja, Indonesia}}

\renewcommand{\headeright}{}

\hypersetup{
pdftitle={Neighbor displacement-based enhanced synthetic oversampling for multiclass imbalanced data},
pdfsubject={q-bio.NC, q-bio.QM},
pdfauthor={I Made Putrama, Peter Martinek},
pdfkeywords={Neighbor, Displacement, Synthetic, Oversampling, Multiclass, Imbalanced Data},
}

\begin{document}
\maketitle

\begin{abstract}
	Imbalanced multiclass datasets pose challenges for machine learning algorithms. These datasets often contain minority classes that are important for accurate prediction. Existing methods still suffer from sparse data and may not accurately represent the original data patterns, leading to noise and poor model performance. A hybrid method called Neighbor Displacement-based Enhanced Synthetic Oversampling (NDESO) is proposed in this paper. This approach uses a displacement strategy for noisy data points, computing the average distance to their neighbors and moving them closer to their centroids. Random oversampling is then performed to achieve dataset balance. Extensive evaluations compare 14 alternatives on nine classifiers across synthetic and 20 real-world datasets with varying imbalance ratios. The results show that our method outperforms its competitors regarding average G-mean score and achieves the lowest statistical mean rank. This highlights its superiority and suitability for addressing data imbalance in practical applications.
\end{abstract}


\section{Introduction}
Imbalanced data remains a persistent challenge in various domains within the enterprise, especially in the context of classification tasks \citep{Patel2020, Cayr2021, Moniz2021, Arafa2022, Ren2023, Tao2023, Yuan2023, Madkour2024}. Imbalanced datasets have a much larger number of majority classes than one or more heavily underrepresented minority classes. This imbalance problem creates substantial challenges for standard machine learning algorithms to accurately identify the minority classes, where correctly classifying these instances is often critical in many applications \citep{Lango2022}. In the financial sector, imbalanced data classification poses significant obstacles, especially in fraud detection applications. Fraudulent transactions occur less frequently than legitimate transactions, resulting in a highly skewed dataset. Failure to accurately detect fraudulent activity can cause major financial losses to organizations \citep{Yu2022a}. Similarly, in manufacturing, as automation advances and production complexity increases, monitoring equipment performance becomes important to prevent failures and ensure safety. This makes collecting equipment data and detecting process patterns and anomalies to support maintenance strategies crucial \citep{Han2020, Bajaj2023}. However, in practice, the equipment failures occur less frequently than normal operations, resulting in imbalanced datasets \citep{Fan2022a, DeGiorgio2023}. Using imbalanced datasets for classification, relying solely on accuracy can be misleading. For example, a dataset with 99\% majority class and 1\% minority class might show a high accuracy of 99\% if the model correctly classifies all majority class instances, despite failing to accurately detect the critical minority class instances \citep{Rezvani2023}.

To deal with imbalanced data, various strategies have been developed for classification tasks involving methods categorized into data level, algorithm level, or a combination of both \citep{Tanha2020, Liu2021a, Niaz2022, Rezvani2023, Ding2023}. These approaches aim to present intelligence diagnostic results in enterprise applications accurately. However, in industrial scenarios, the complexity of data acquisition under the aforementioned conditions often leads to data collection that may exhibit small-sample problems, making it challenging to establish highly accurate models when insufficient data is available \citep{Pan2022}. Additionally, the data may contain errors or anomalies, where noise in imbalanced data significantly affects the performance of the diagnostic algorithm \citep{Patange2023, Pancaldi2023}. Moreover, while the use of existing oversampling, undersampling, or hybrid methods has been found effective in dealing with imbalanced data, this approach does not always produce synthetic samples that are entirely free of noise or avoid overlapping between data. This is especially prominent for data that naturally has many overlapping data points. In such conditions, existing methods that generate new samples based on the unclean original data distribution tend to produce increasingly noisy samples, causing the generated data to deviate significantly from the original pattern, which ultimately affects the overall accuracy and effectiveness of the model.

In this paper, we introduce a hybrid resampling method called Neighbor Displacement-based Enhanced Synthetic Oversampling (NDESO), which aims to correct the noisy data points within each class by moving them closer to their centroids before oversampling is performed to balance the data distribution. This method identifies data points that are located around $k$ neighbors with different classes. The displacement is done by analyzing the distance of the relative position of the data point to its neighbors and then moving it closer to the corresponding class centroid based on the distance. This displacement process adjusts the data points towards their corresponding centroid, aligning them more closely with the characteristics of their class, while preserving the original class label. With the adjusted position of noisy data, class balancing is then performed through oversampling minority classes, ensuring improved outcomes while preserving the patterns of the original data. This method was thoroughly evaluated through a comparative study against 14 distinct resampling methods in an experiment that utilized nine different classifiers across 20 real-world datasets. Statistical tests indicate that our proposed method yields significantly smaller critical differences, demonstrating a notable improvement in performance compared to other methods.

The remainder of this manuscript is structured as follows: Section~\ref{sec: related} explores existing research related to the field. Section~\ref{sec: objectives} outlines the objectives of our study. Section~\ref{sec: method} describes the method used for our proposed framework. Experiments covering data collection, system specifications, testing procedures, and results are described in Section~\ref{sec: experiment} which thoroughly demonstrate the efficacy of our approach. Section~\ref{sec: discussion} provides a summary of our findings and offers suggestions for future research directions. Finally, Section~\ref{sec: conclusion} provides the conclusion of our work.

\section{Related works}\label{sec: related}
Real-world classification datasets often have imbalanced class distributions, with many samples in one class and few in others. Many datasets feature multiclass or multilabel distributions, involving more than two classes or allowing instances to belong to multiple classes. Existing approaches can be grouped into data-level, algorithmic, or mixed methods that combine both approaches to address this imbalanced data problem. However, in this study, our approach is oriented towards data-level methods.

The data-level method uses a re-sampling strategy: under-sampling, over-sampling, or hybrid-sampling. Random undersampling (RUS) and random oversampling (ROS) are two techniques used to remove or add samples to make the classes balanced \citep{Wang2024a, Vairetti2024}. RUS is the simplest undersampling technique, where the existing majority of samples are removed until the class distribution is balanced. Another widely used undersampling technique is NearMiss, introduced by \citep{Zhang2003}, which seeks to address class imbalance by selectively removing samples from the majority class. NearMiss eliminates majority class instances to enhance their separation when it detects proximity between instances of different classes. However, undersampling may inadvertently lead to reduced information, given the potential for eliminating valuable class instances \citep{DeGiorgio2023}. Consequently, various methods have been proposed to improve data cleaning by identifying and removing redundant patterns and noise in the dataset, thereby enhancing classifier performance. EditedNearestNeighbors (ENN), Condensed Nearest Neighbors (CNN), and Tomeklinks are several methods based on this technique \citep{Kulkarni2020}. These methods remove instances near the decision boundary between classes, enhancing class separation and improving classifier performance.

Unlike RUS, which removes example data, ROS replicates minority samples randomly until the class distribution is balanced. As a result, this method has been criticized for adding more data without contributing new information, which can lead to overfitting. To handle this problem, a well-known approach called Synthetic Minority Over-sampling TEchnique (SMOTE) was proposed by \citep{Chawla2002}. SMOTE creates syntactic samples rather than directly copying from minority classes. It generates additional instances through interpolation within the $k$-Nearest Neighbors of a minority class \citep{Yuan2023}. Initially, a sample (x) is selected from the minority class, followed by another sample (y) from its $k$-Nearest Neighbors set within the same class. Subsequently, the new sample $x'$ is created through linear interpolation using the following formula: {\footnotesize \begin{align}
		x' = x + rand(0, 1) + |x-y| \nonumber
\end{align}}The iterative process of creating new synthetic samples is carried out until the desired amount of oversampling is achieved. Despite its straightforward and innovative approach, SMOTE has a key shortcoming called overgeneralization, as it ignores within-class imbalance, which tends to increase class overlap. Furthermore, the generation of new synthetic points to address the sparsity of the minority class may inadvertently introduce uninformative and noisy instances that do not accurately represent the true underlying data patterns \citep{Wang2023d}. These extraneous patterns can contribute to overfitting, potentially reducing the model's generalizability. Since then, there have been other variations to the SMOTE approach, including Borderline-SMOTE, KMeans-SMOTE, and SVM-SMOTE \citep{Ahsan2024}. Other variants exist that combine noise removal techniques aiming to clean data, such as SMOTE-Tomek and SMOTE-ENN, which are methods designed to oversample minority classes in a dataset while also cleaning noisy instances \citep{DeGiorgio2023}. These versions of the SMOTE technique focus on samples within the boundary area between classes, creating synthetic samples exclusively on the dividing line of two classes to avoid overfitting. Another recent hybrid variant was proposed by \citep{Wang2023d}, which uses SMOTE for oversampling and Edited Displacement-based $k$-Nearest Neighbors (ECDNN) for undersampling, known as SMOTE-CDNN. The solution aims to minimize the within-cluster distance problem by using centroid displacement for class probability estimation. Other techniques similar to SMOTE, including Adaptive Synthetic Sampling (ADASYN) and Stacking algorithms, have been proposed to address multiclass imbalance problems, beating previous techniques regarding accuracy and sensitivity. ADASYN was introduced by \citep{He2008} and focused on generating new samples by prioritizing instances based on the density distribution of the minority class. It creates more samples in areas where the minority class density is low and fewer samples where the minority class density is higher.	
\begin{figure}[!htp]
	\centering
	\begin{minipage}[b]{.45\linewidth}
		\includegraphics[page=1,width=\linewidth]{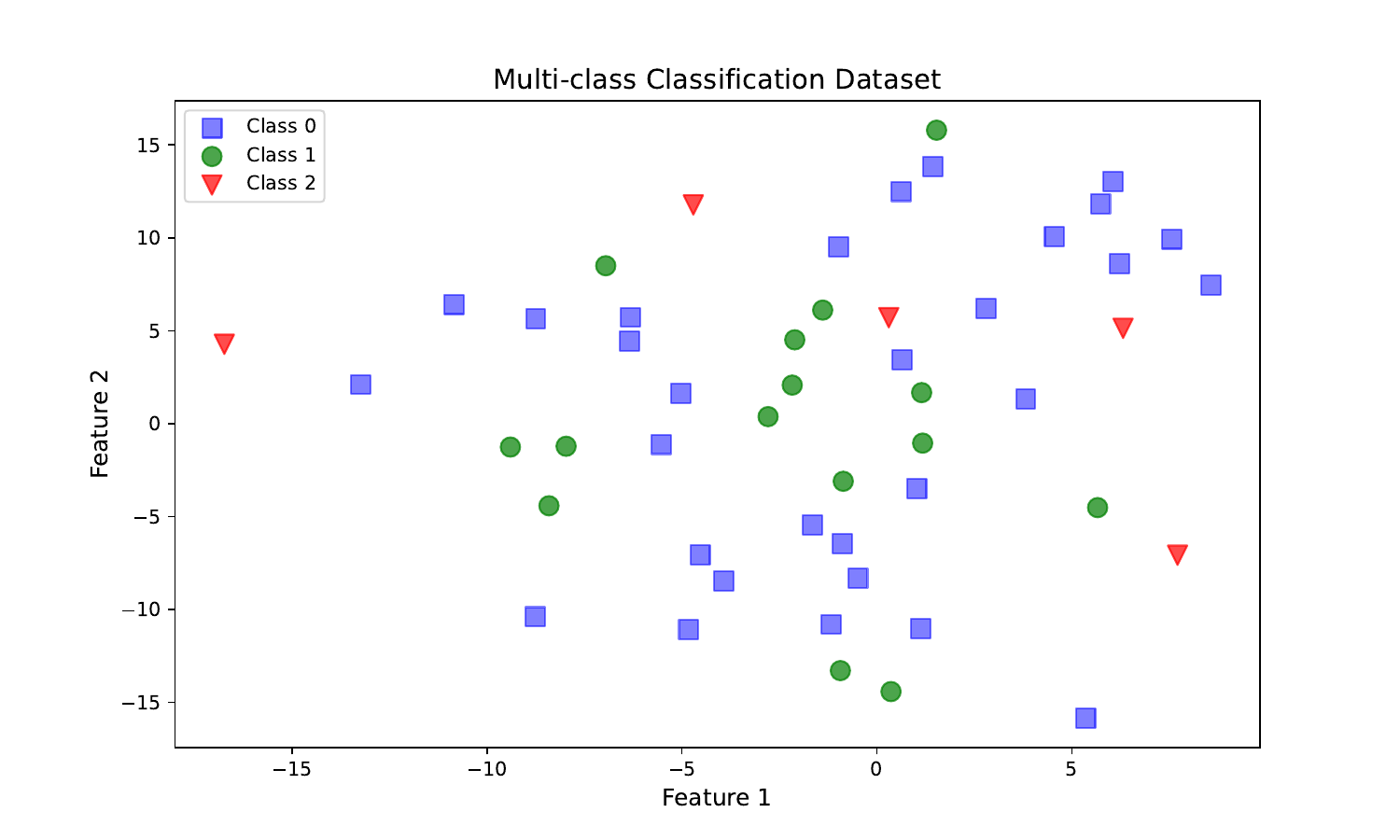}
		\centering (a)		
	\end{minipage}
	\begin{minipage}[b]{.45\linewidth}
		\includegraphics[page=1,width=\linewidth]{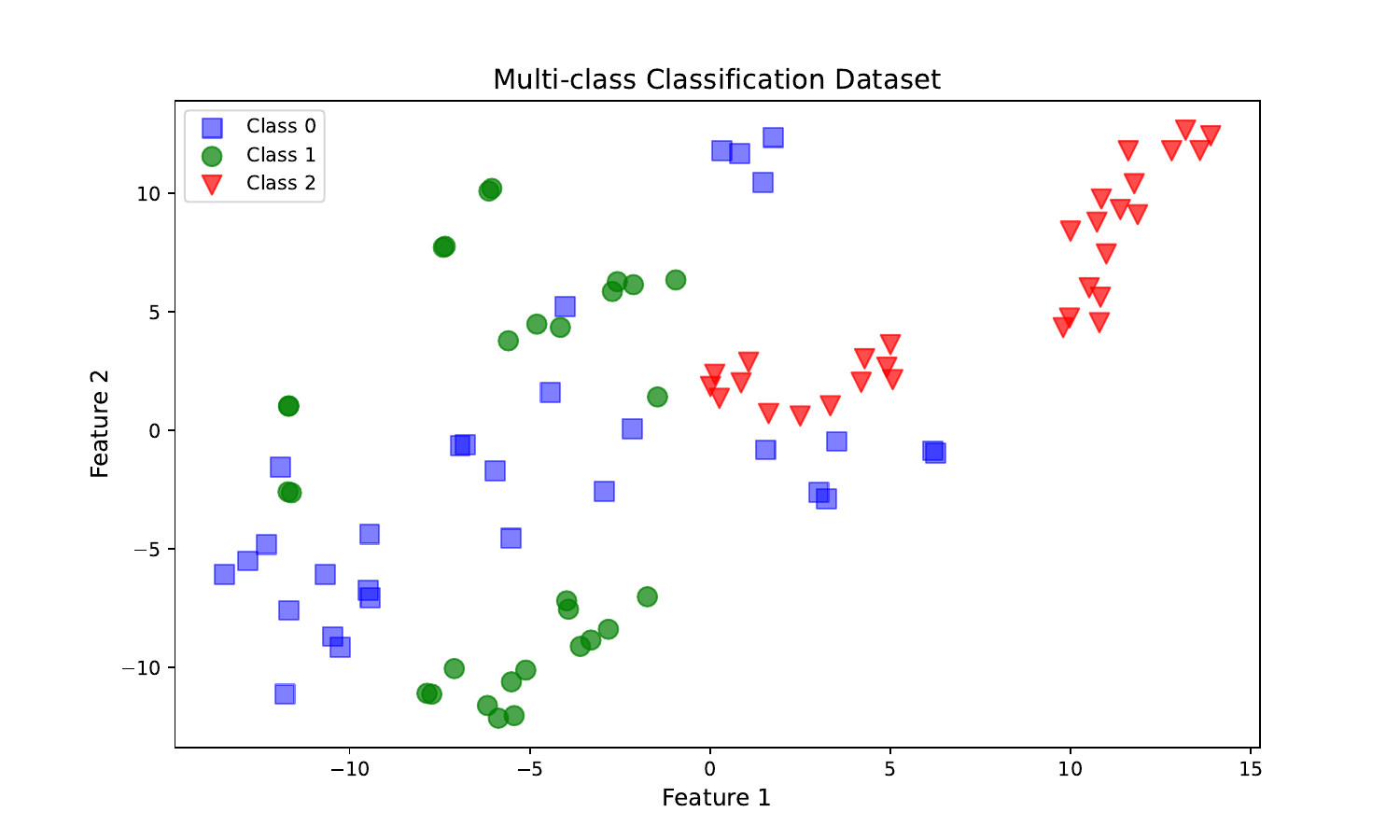}
		\centering (b)		
	\end{minipage}
	\caption{Visualization of resampling on a sparse multiclass dataset: (a) original dataset; (b) noisy resampled dataset using SMOTE}
	\label{fig:sparse-data}
\end{figure}

Despite the efficacy demonstrated by the earlier methods, our observations indicate that these techniques remain vulnerable to sparse, overlapping data points of multiclass imbalanced datasets. This leads to issues such as 'within class imbalance' and 'small disjunct problem,' which have also been emphasized by \citep{Liu2021a, Islam2022, Yuan2023}. When the minority classes have few samples, and oversampling is performed within clusters, the majority class overwhelmingly dominates the dataset, leaving the remaining classes with only sparse instances. Furthermore, employing an approach like SMOTE, which generates synthetic instances by interpolating between current samples of the minority class, may not be effective in sparse datasets, as shown in Fig.~\ref{fig:sparse-data}. This is because the minority class samples are scattered and limited but contain overlapping instances that belong to different classes, potentially leading to synthetic samples that do not accurately represent realistic or potential minority class samples. Consequently, this can result in a poorly performing model on new, unseen data. Moreover, it has been observed that resampling techniques like ADASYN and KMeans-SMOTE do not always consistently succeed in identifying enough samples that form clusters. They also often struggle to determine the minimum number of neighbors required for some datasets, which could affect the efficiency of these methods. These challenges motivated our study to explore innovative and stronger strategies for dealing with sparse datasets. 
\begin{figure}[!htp]
	\centering
	\begin{minipage}[b]{.35\linewidth}
		\includegraphics[page=1,width=\linewidth]{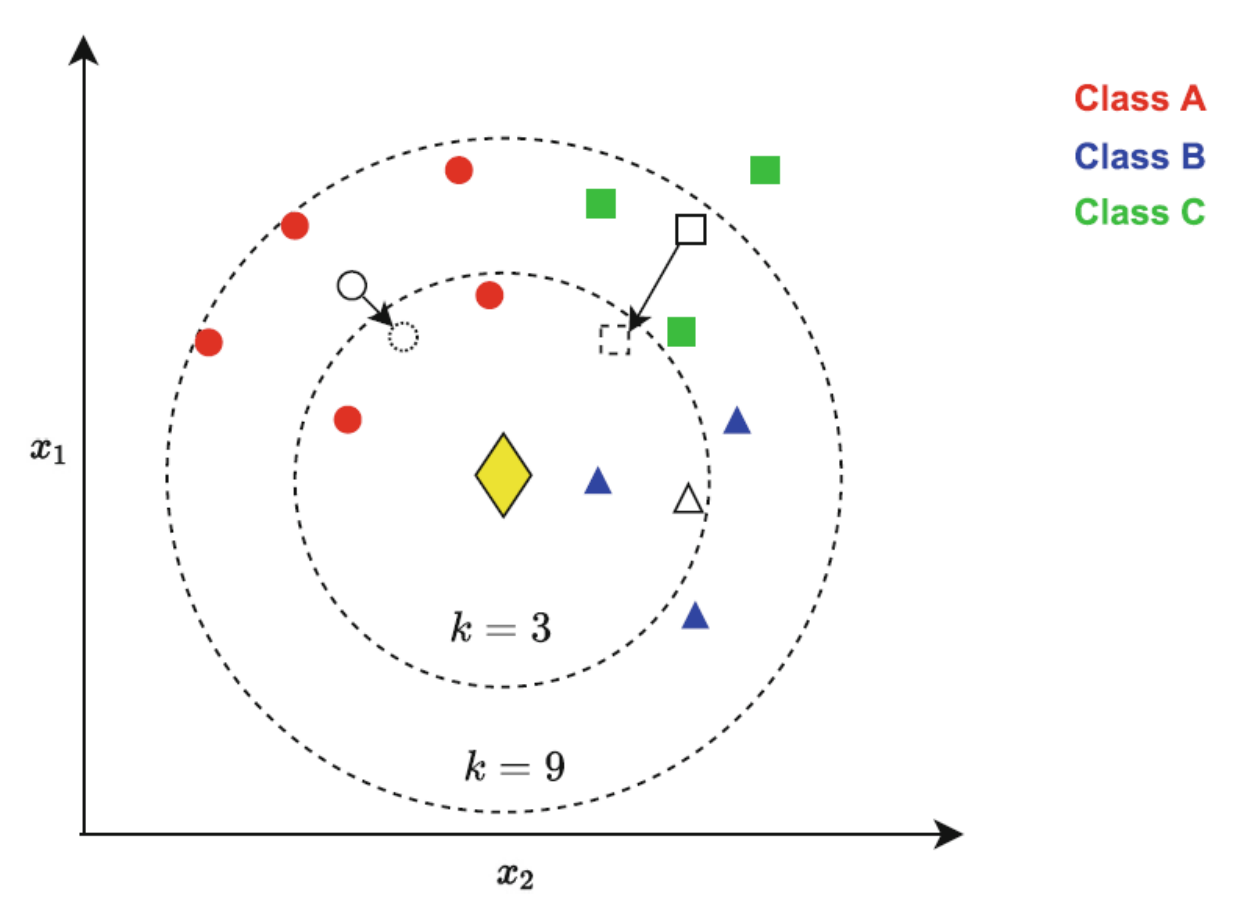}
		\vspace{.1pt}
	\end{minipage}
	\caption{Visual illustration of the CDNN algorithm \citep{Wang2022d}}
	\label{fig: cdnn}
\end{figure}

Our method differs from the ECDNN approach proposed by \citep{Wang2022d}, which determines a test data point's label based on the minimum displacement of the nearest class centroid of its $k$-neighbors after incorporating the test instance into the set. As illustrated in Fig.~\ref{fig: cdnn}, a test example (yellow diamond) can be assigned to class B since it produces the smallest displacement of the class centroid compared to other classes. The approach is employed for undersampling to eliminate noisy data points when their label are predicted differently. It is integrated with SMOTE to balance the dataset, forming a technique known as SMOTE-CDNN \citep{Wang2023d}. Although this approach successfully balances the dataset, in cases involving sparse multiclass imbalanced data where minority samples hold useful information, altering their labels or removing them when their predicted class differs from their original class could unintentionally eliminate valuable instances, potentially affecting classification performance in real-world scenarios. Therefore, instead of discarding these noisy data points, our method employs a unique strategy that adjusts their positions by analyzing the distances between a data point and its $k$-neighbors. It displaces the position only if the data point belongs to the minority class within its neighborhood (i.e., if the majority of its neighbors belong to a different class). We hypothesize that this adjustment will preserve their class labels and characteristics while mitigating their influence as noise. This refinement is expected to enhance class separation and promote a more balanced representation, distinguishing our approach from existing methodologies.		
\section{Objectives}\label{sec: objectives}
\begin{figure}[!htp]
	\centering
	\begin{minipage}[b]{.85\linewidth}
		\includegraphics[page=1,width=\linewidth]{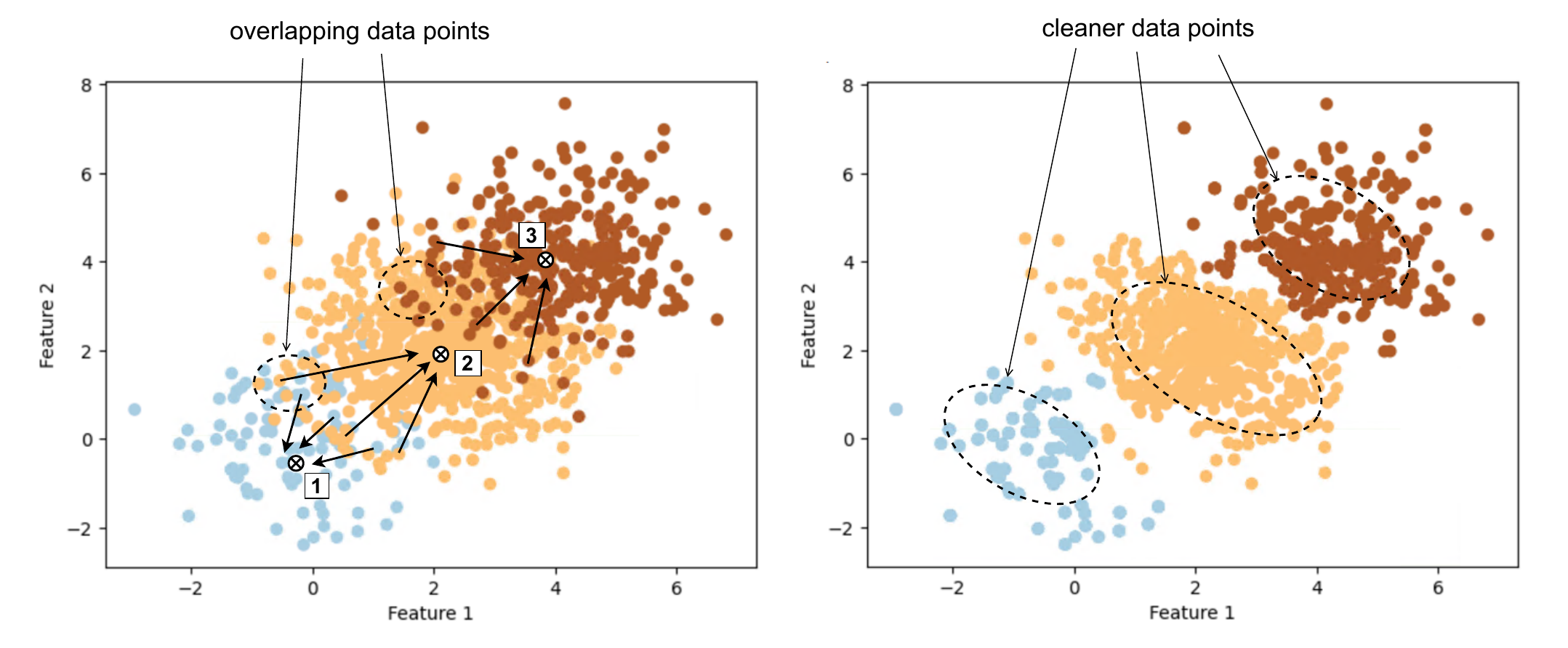}
		\vspace{.1pt}
	\end{minipage}
	\caption{The overlapping (before) and cleaned (after) data points}
	\label{fig: cleaning}
\end{figure}

The primary objective of this study is to evaluate the effectiveness of displacing noisy data points towards the center of their respective classes, as depicted in Fig.~\ref{fig: cleaning} on the left, resulting in a clearer separation illustrated on the right before additional samples are generated to balance the dataset. This approach aims to yield a more accurate representation of the underlying patterns within the data. While our approach utilizes random oversampling as a straightforward technique to achieve optimal performance, by using cleaner, centroid-aligned data points as a base will mitigate the common issue of overfitting associated with random oversampling. Specifically, we seek to achieve the following objectives:
\begin{enumerate}
	\item Create a resampling method that refines imbalanced datasets by shifting noisy data points towards their corresponding centroids, thus improving the dataset's quality. 
	\item Integrate random oversampling to achieve class balance using the cleaned data points, thereby reducing computational overhead while enhancing model performance and preventing underfitting.
	\item Conduct a comprehensive comparison of the proposed method with baseline resampling methods using multiple classifiers to evaluate the effectiveness and robustness of the approach.
	\item Assess the performance of the resampling methods using the various metrics to ensure balanced classification performance across imbalanced datasets and apply the Friedman-Nemenyi non-parametric statistical tests to validate the results.
\end{enumerate}
\section{Method}\label{sec: method}
This section outlines the principle of our approach to perform the \textit{displacement of the noisy data points} (NDE in Section~\ref{sec: nde}) before performing \textit{random oversampling} (NDESO in Section~\ref{sec: ndeso}).	
\begin{algorithm}[!htp]
	\caption{Random oversampling with noisy data points displacement}
	\label{algo: ndeso}
	\begin{algorithmic}[1]
		\Require data $X$, label $y$, and neighbors $k$
		\Procedure{nde}{X, y, cdist='eucledian', k=5}
		\State $d = \text{cdist}(X, X)$ \Comment{calculate pairwise distance between all points}
		\State $\text{idxs} = \text{sort}(d)[:, 1:k + 1]$ \Comment{get k nearest neighbors' indices}
		\State $\text{disps} = \text{set}()$
		\For{$i \in \text{index}(X)$}
		\State $\text{nbs} = \text{idxs}[i]$
		\State $\text{same} = \text{sum}(y[\text{nbs}=y[i]])$
		\State $\text{diff} = k - \text{same}$
		\If{\text{diff} > \text{same}}
		\State $\text{disps}.add(i)$
		\State $\text{disps}.update(\text{nbs}[y[\text{nbs}]=y[i]])$
		\EndIf
		\EndFor
		\State $R = {X[y=cls]\,\text{for}\,cls \in \text{classes}}$
		\For{$i \in \text{disps}$} \Comment{displace the data points}
		\State $r_{c_i} = R[i]$
		\State $S_i = \text{cdist}([X[i]], [r_{c_i}])$
		\State $\vec{S_v} = (r_{c_i} - X[i]) / S_i$
		\State $\phi_i = mean(d[i, \text{idxs}[i]])$
		\State $X[i] = r_{c_i} - \vec{S_v} * \phi_i$
		\EndFor
		\State \Return $X, y$
		\EndProcedure
		\Procedure{ndeso}{X, y, cdist='eucledian', k=5}
		\State $X, y = NDE(X, y, cdist, k)$
		\State \Call{RandomOver}{X, y} \Comment{perform random oversampling}
		\EndProcedure
	\end{algorithmic}
\end{algorithm}	
\subsection{Neighbor-based displacement enhancement (NDE)}\label{sec: nde}
The foundational version of our algorithm, known as Neighbor-based Displacement Enhancement (NDE), is designed to address noisy class data points by strategically repositioning them closer to their respective centroids to enhance class separation, which operates as follows. Given a set of data points $X = \{(x_i, c_i):x_i \in \mathbb{R}^n, c_i \in C\}$, where $x_i$ is a $n$-dimensional vector and $c_i$ is its corresponding class label from the set of class labels $C$, suppose there are data points $x_i$ associated with different class labels $c_i$ that overlap each other, as depicted in Fig.~\ref{fig: movables}. In order to displace the overlapping data points, the following steps are followed:
\begin{figure}[!htp]
	\centering
	\begin{minipage}[b]{.75\linewidth}
		\includegraphics[page=1,width=\linewidth]{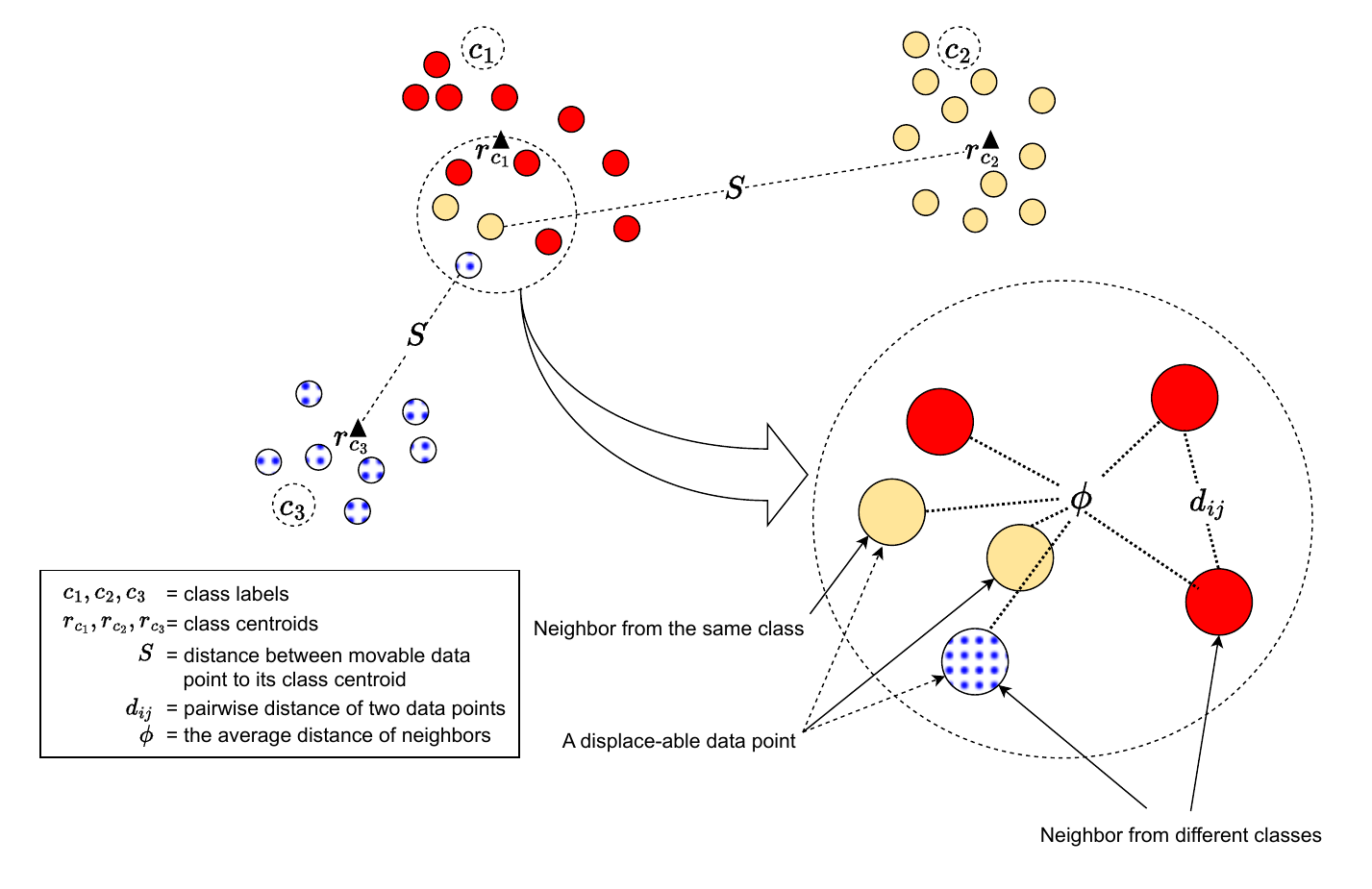}
		\vspace{.1pt}
	\end{minipage}
	\caption{Visual illustration of displace-able data point identification}
	\label{fig: movables}
\end{figure}
\begin{enumerate}
	\item A pairwise distance matrix $D = [d_{ij}]$ is computed for each pair of data points $(x_i, x_j) \in X \times X$, where $X$ is the dataset.		
	\item Based on the distance matrix $D$, a set of indices $I$ is obtained for each data point, containing the indexes of its $k$ nearest neighbors
	\item Iterate over the dataset $X$ to identify a set of displaceable data points $M$, where a data point $x_i$ is deemed a candidate for displacement if it satisfies the following condition:
	{\noindent\footnotesize
		\begin{align}
			A &= \left| \{x_j \in \mathcal{N}_k(x_i) \mid c_j = c_i \} \right|, \nonumber \\
			B &= \left| \{x_j \in \mathcal{N}_k(x_i) \mid c_j \neq c_i \} \right|, \nonumber \\
			M &= \{ x_i \in X \mid A < B \}
		\end{align}
	}
	where:
	\begin{itemize}
		\item $\mathcal{N}_k(x_i)$ represents the set of $k$ nearest neighbors of $x_i$
		\item $c_i$ is the class label of the data point $x_i$
		\item $c_j$ is the class label of the neighbor $x_j$
	\end{itemize}	
	\item Compute the set of centroids $R$:
	{\noindent\footnotesize
		\begin{align}
			R = \{ r_{c_i} : r_{c_i} = \frac{1}{|C_{c_i}|} \sum_{x_j \in C_{c_i}} x_j \}
	\end{align}} where $r_{c_i}$ is the centroid of class $c_i$ and $C_{c_i}$ is the set of data points belonging to class $c_i$.		
	\item Iterate through the set of displace-able data points $M$ to perform the following:
	\begin{enumerate}
		\item Get the centroid $r_{c_i}$ for a displace-able data point $x_i$
		\item Compute the normalized direction vector from $x_i$ to the centroid $r_{c_i}$
		{\noindent\footnotesize
			\begin{align}
				S_i = \left\| x_i - r_{c_i} \right\| \qquad \vec{S_v} = \frac{(r_{c_i} - x_i)}{S_i}
		\end{align}}
		\item Compute the displacement distance $\phi_i$ for the data point $x_i$:
		{\noindent\footnotesize
			\begin{align}
				\phi_i = \frac{1}{k}\sum_{x_j \in \mathcal{N}_k(x_i)}^{} d_{ij}
		\end{align}}
		\item Displace $x_i$ closer to its centroid as $x'_i$:
		{\noindent\footnotesize
			\begin{align}
				x'_i = r_{c_i} - \vec{S_v} * \phi_i
		\end{align}}
	\end{enumerate}		
\end{enumerate}
This NDE algorithm uses a default of $k=5$ neighbors and the Eucledian distance metric.

\subsection{Neighbor-based displacement enhanced synthetic oversampling (NDESO)}\label{sec: ndeso}
To address class imbalance, random oversampling is applied to the dataset after removing noisy data points. This enhanced version of the NDE algorithm is called Neighbor-based Displacement Enhanced Synthetic Oversampling (NDESO). This oversampling process utilizes the \textit{RandomOverSampler} method from the \textit{imblearn}\footnote{\url{https://imbalanced-learn.org}} library. In this technique, new data points are generated by randomly duplicating samples from the minority class. If $X_{\text{minority}}$ represents the set of data points from the minority class, the oversampling process creates additional data points by randomly selecting from $X_{\text{minority}}$ until the number of minority class points is equal to that of the majority class, denoted as $X_{\text{majority}}$, as given in the following formula:

{\noindent\footnotesize
	\begin{align}
		X_{new} &= \text{random\_sample}\,(X_{\text{minority}}, |X_{\text{majority}}| - |X_{\text{minority}}|) \nonumber \\
		X'_{\text{minority}} &= X_{\text{minority}} \cup X_{new}
\end{align}}
By cleaning the dataset in advance, this approach seeks to significantly improve the quality of the resampled data points, refining the oversampling process while minimizing additional overhead for a more effective class balance.

The complete algorithm for the procedures of our proposed approach is given in Algorithm~\ref{algo: ndeso}.

\subsection{Performance metrics}
\subsubsection{G-mean}
We calculate precision, recall, F1-score, and G-mean as evaluation metrics during the experiment. However, we have chosen to report G-means in this paper due to space constraints. For a complete proof of the testing output, please refer to the GitHub repository link provided at the end of this manuscript. 

Geometric mean (G-mean) is a widely recognized performance metric for imbalanced classification tasks, aiming to achieve a balanced assessment across various classes by focusing on both positive and negative classes. Mathematically, G-mean is defined as follows \citep{Jia2024}:

{\footnotesize \begin{flalign}
		\textit{G-mean} = \sqrt{Sensitivity \times Specificity}
\end{flalign}}
where:
\begin{itemize}
	\item \textit{Sensitivity(Recall)}: The ratio of the actual positive instances that are correctly classified by the classifier. Sensitivity is often denoted as True Positive Rate (TPR).
	{\footnotesize
		\begin{flalign}
			Sensitivity\,(TPR) = \frac{True\,Positives}{True\,Positives + False\,Negatives}
	\end{flalign}}
	\item \textit{Specificity}: The ratio of the actual negative instances that are correctly classified by the classifier. Specificity is often denoted as True Negative Rate (TNR).
	{\footnotesize
		\begin{flalign}
			Specificity\,(TNR) = \frac{True\,Negatives}{True\,Negatives + False\,Positives}
	\end{flalign}}
\end{itemize}
The G-mean score typically ranges from 0 to 1. A score of 1 indicates a perfect balance between Sensitivity (Recall) and Specificity (True Negative Rate), meaning the model performs equally well on both positive and negative classes. A score closer to 0 indicates poorer performance in achieving this balance. In practice, the G-mean value varies for imbalanced classification tasks. However, values closer to 1 are targeted to indicate a better trade-off between Sensitivity and Specificity.
\subsubsection{Friedman and post-hoc Nemenyi tests}
In this study, we compared our resampling method with several baseline techniques. To evaluate performance, we used statistical tests, in particular the Friedman and Nemenyi tests, which are recommended and widely used for distinguishing statistically significant differences between approaches, to ensure stable results against randomness in experimental analysis \citep{Pereira2015, Ma2022a, M2024, Jia2024, Din2024}.

The Friedman and Nemenyi tests are nonparametric statistical methods used to assess the performance differences among multiple resampling methods in classification experiments. These tests do not assume any specific data distribution. The \textit{null} hypothesis implies that all methods perform similarly across the tested datasets while rejecting this hypothesis (with a significance level $\alpha$ of 0.05) indicates significant performance differences among the resampling methods \citep{Demsar2006}.

For each dataset $i \in D$, the resampling methods are ranked from best to worst as 1 to $k$, where $k$ denotes the number of resampling methods, including our proposed approach. The mean rank $R_j$ for the $j$-th resampling method is defined by:

{\footnotesize \begin{flalign}
		R_j = \frac{1}{N} \sum_{i \in D} r_i^j
\end{flalign}}
where $r_i^j$ represents the rank of the $j$-th resampling method on the $i$-th dataset, and $N$ is the total number of datasets $D$. The Friedman test is then defined as:
{\footnotesize \begin{flalign}
		\chi^2_{\text{Friedman}} = \frac{12N}{k \cdot (k+1)} \left[ \sum_{j=1}^{k} \left( R_j^2 \right) - \frac{k \cdot (k+1)^2}{4} \right]
\end{flalign}}

Suppose the \textit{null} hypothesis is rejected by the Friedman test. In that case, we proceed with the post hoc Nemenyi test for a pairwise comparison among the resampling method based on a critical difference (CD), which is defined as:
{\footnotesize \begin{flalign}
		CD = q_{\alpha} \cdot \sqrt{\frac{k \cdot (k+1)}{6 \cdot N}}
\end{flalign}}
where $q_{\alpha}$ is the critical value from the Studentized range distribution for a given significance level $\alpha$, based on the CD value, resampling methods whose mean ranks differ by at least the CD value are considered significantly different. Likewise, if the difference in mean rank between the two resampling methods is smaller than the CD value, then the difference in performance is considered not statistically significant. In other words, they are statistically comparable.

\begin{table*}[!htp]
	\caption{Datasets description}\label{tbl: real-data}
	\fontsize{7pt}{6pt}\selectfont
	\renewcommand{\arraystretch}{1.6}
	\begin{minipage}{1\textwidth}
		\setlength{\tabcolsep}{6pt}
		\begin{tabular}[t]{L{0.1\linewidth}L{0.15\linewidth}C{0.08\linewidth}C{0.06\linewidth}C{0.05\linewidth}C{0.05\linewidth}C{0.2\linewidth}C{0.11\linewidth}}
			\toprule
			\textbf{Dataset} & \textbf{Description} & \textbf{Features} & \textbf{Instances} & \textbf{Classes}  & \textbf{Majors} & \textbf{Minors} & \textbf{Imbalance Ratio}\\
			\midrule
			autos$\clubsuit$ & Automobile & 25 & 159 & 6 & 48 & [46, 29, 20, 13, 3] & 16.0\\
			balance & Psychological experimental & 4 & 625 & 3 & 288 & [288, 49] & 5.88 \\
			contraceptive & Contraceptive method choice & 9 & 1,473 & 3 & 629 & [511, 333] & 1.89 \\
			dermatology & Dermatology patients  & 34 & 366 & 6 & 112 & [72, 61, 52, 49, 20] & 5.60\\
			ecoli$\clubsuit$ & Protein localization & 7 & 336 & 9 & 139 & [77, 52, 35, 20, 5, 4, 2, 2] & 69.50 \\
			glass$\clubsuit$ & Glass chemical properties & 9 & 214 & 6 & 76 & [70, 29, 17, 13, 9] & 8.44 \\
			hayes-roth & People characteristics & 4 & 132 & 3 & 51 & [51, 30] & 1.70 \\
			lymphography$\clubsuit$ & Patient's radiological examination & 18 & 148 & 4 & 81 & [61, 4, 2] & 40.50 \\
			new-thyroid & Thyroid diseases detection & 5 & 215 & 3 & 150 & [35, 30] & 5.0 \\
			pageblocks$\clubsuit$ & Document page layout blocks & 10 & 548 & 5 & 492 & [33, 12, 8, 3] & 164.0 \\
			penbased & Pen-Based recognition of handwritten digits & 16 & 1,100 & 10 & 115 & [115, 114, 114, 114, 106, 106, 106, 105, 105] & 1.10 \\
			segment & Image segmentation & 18 & 2,310 & 7 & 330 & 330, 330, 330, 330, 330, 330] & 1.0 \\
			shuttle$\clubsuit$ & Space shuttle dataset of the Statlog project & 9 & 2,175 & 5 & 1,706 & [338, 123, 6, 2] & 853.0 \\
			svmguide2 & Benchmark dataset & 20 & 391 & 3 & 221 & [117, 53] & 4.17 \\
			svmguide4 & Benchmark dataset & 10 & 300 & 6 & 56 & [56, 53, 47, 44, 44] & 1.27 \\
			thyroid$\clubsuit$ & Thyroid diseases detection & 21 & 720 & 3 & 666 & [37, 17] & 39.18 \\
			vehicle & Vehicle object detection & 18 & 846 & 4 & 218 & [217, 212, 199] & 1.10 \\
			vowel & Vowel recognition of British English & 10 & 528 & 11 & 48 & [48, 48, 48, 48, 48, 48, 48, 48, 48, 48] & 1.0 \\
			wine & Chemical characteristics of wines  & 13 & 178 & 3 & 71 & [59, 48] & 1.48 \\
			yeast$\clubsuit$ & Protein localization sites in yeast cells & 8 & 1,484 & 10 & 463 & [429, 244, 163, 51, 44, 35, 30, 20, 5] & 92.60 \\
			\bottomrule
		\end{tabular}
		\vspace{-10pt}
		\begin{tablenotes}
			\begin{multicols}{1}
				\item[1][$\clubsuit$] dataset used in NDE testing
			\end{multicols}
		\end{tablenotes}
	\end{minipage}	
\end{table*}	

\section{Experiment}\label{sec: experiment}
\subsection{Datasets and environment}

\subsubsection{Synthetic datasets}
We utilize a synthetic dataset to provide a clear visualization of the data before and after resampling, using our algorithm and comparing it with others. This dataset consists of three distinct classes (class 0, class 1, class 2), generated based on a normal distribution, with a ratio of 50:500:100. To introduce some overlap between the classes, random noise was added to the data points, with the noise generated from a normal distribution and scaled by a factor of 0.75.

\subsubsection{Real-world datasets}
The real-world datasets are sourced primarily from Knowledge Extraction\footnote{\url{https://sci2s.ugr.es/keel/imbalanced.php}} and OpenML\footnote{\url{https://www.openml.org/search?type=data}}. Details of the real-world multiclass dataset used in the experiments are presented in Table~\ref{tbl: real-data}, with the number of classes ranging from 3-11 and the imbalance ratio ranging from 1.0 to 853.0.

\subsubsection{System environment and libraries}\label{sec: specs}
Throughout the experiment, Python running on a Virtual OS with Linux Server was primarily used, hosted on hardware featuring 32GB of RAM, an Intel i9 processor with 16 cores, and a 25GB SSD hard drive. We evaluated our proposed method against several baseline resampling techniques across various classifiers. The samplers and classifiers were implemented using the scikit-learn library\footnote{\url{https://scikit-learn.org}}, with some utilizing the imbalanced-learn library\footnote{\url{https://imbalanced-learn.org}}. Those resamplers are detailed in Section~\ref{sec: related}, while classifiers include Support Vector Machine (SVC), $k$-Nearest Neighbors ($k$-NN), Decision Tree, Random Forest, Multi-layer Perceptron (MLP), Easy Ensemble Classifier, RUS Boost Classifier, Balanced Bagging Classifier, and Balanced Random Forest Classifier, each offering unique learning paradigms and handling imbalanced datasets differently. SVC and $k$-NN are instance-based methods sensitive to data distribution, making them suitable for assessing changes in class separation. DecisionTree and RandomForest are tree-based models that excel at capturing non-linear relationships, with RandomForest offering sophisticated ensemble learning capabilities. MLP, a neural network model, tests the performance of algorithms requiring balanced data for gradient-based optimization. Easy Ensemble, RUS Boost classifiers, Balanced Bagging, and Balanced Random Forest, as ensemble techniques designed for imbalanced data, help evaluate the ability of our resampling technique to mitigate class imbalance across varying classifier characteristics.

\begin{figure*}[!htp]
	\centering
	\graphicspath{ {figures/nde} }
	\begin{minipage}{1\textwidth}
		\begin{subfigure}[t]{1\textwidth}
			\centering
			\includegraphics[page=1,width=\linewidth]{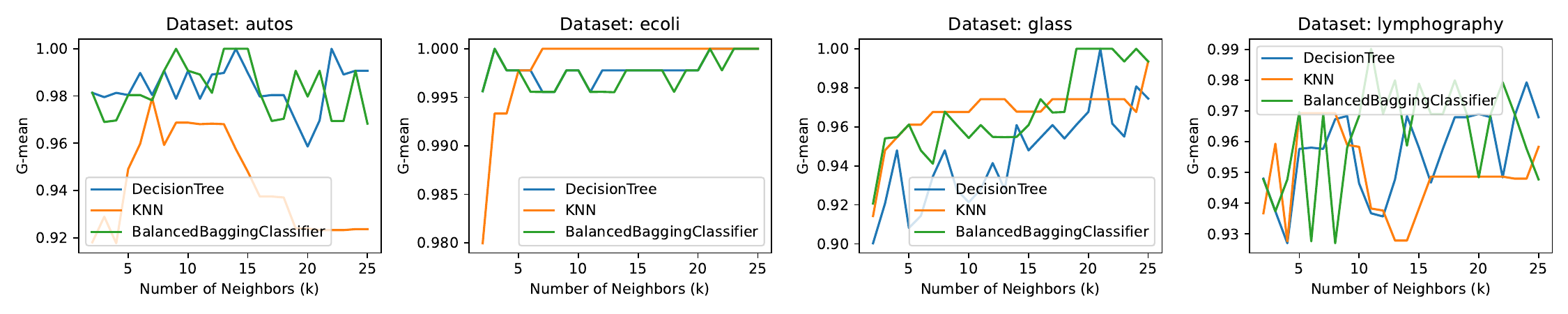}
		\end{subfigure}
	\end{minipage}
	\begin{minipage}{1\textwidth}
		\begin{subfigure}[t]{1\textwidth}
			\centering
			\includegraphics[page=2,width=\linewidth]{neighbors_results}
		\end{subfigure}
	\end{minipage}
	\caption{G-mean scores of our NDE algorithm evaluated across three classifiers for various $k$-nearest neighbors (2-25)}
	\label{fig: test_k}
\end{figure*}

\begin{figure*}[!t]
	\centering
	\graphicspath{ {figures/nde} }
	\begin{minipage}{1\textwidth}
		\begin{subfigure}[t]{1\textwidth}
			\centering
			\includegraphics[page=1,width=\linewidth]{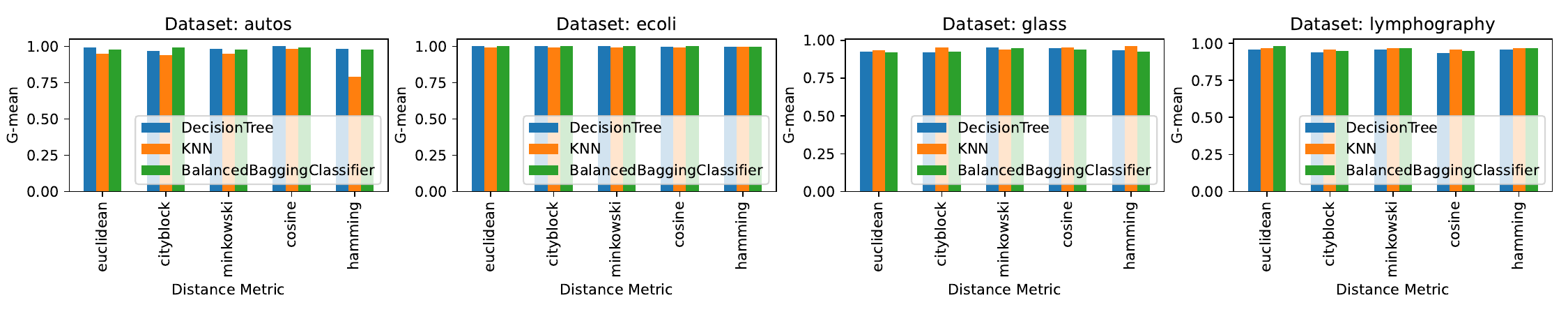}
		\end{subfigure}
	\end{minipage}
	\begin{minipage}{1\textwidth}
		\begin{subfigure}[t]{1\textwidth}
			\centering
			\includegraphics[page=2,width=\linewidth]{metric_results}
		\end{subfigure}
	\end{minipage}
	\caption{G-mean scores of our NDE algorithm evaluated across three classifiers for various distance metrics}
	\label{fig: test_metrics}
\end{figure*}	

\begin{figure*}[!htp]
	\centering
	\graphicspath{ {figures/nde} }
	\begin{minipage}{1\textwidth}
		\begin{subfigure}[t]{1\textwidth}
			\centering
			\includegraphics[page=1,width=\linewidth]{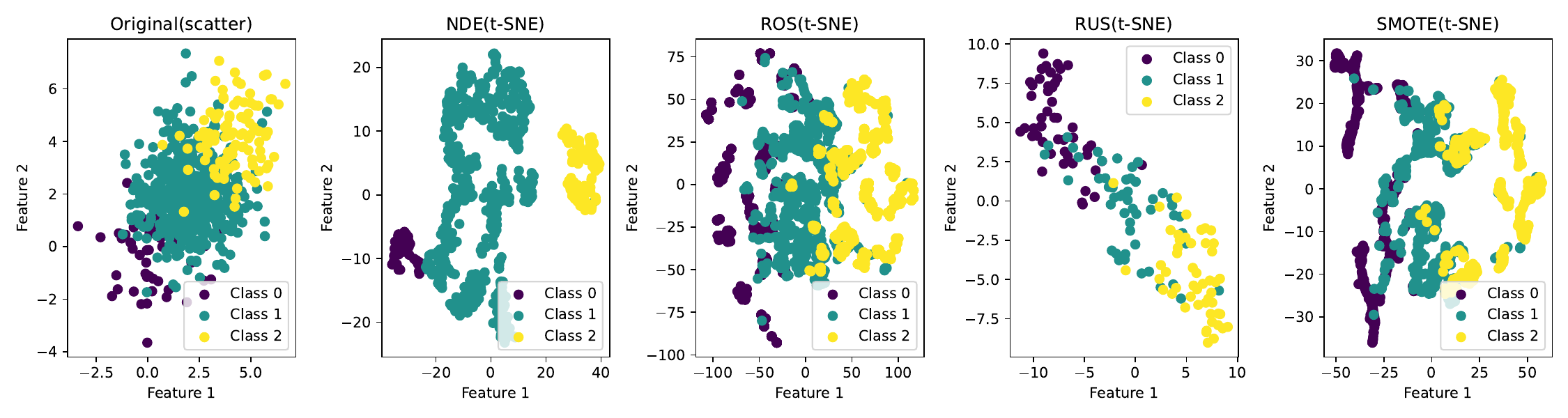}
		\end{subfigure}
	\end{minipage}
	\begin{minipage}{1\textwidth}
		\begin{subfigure}[t]{1\textwidth}
			\centering
			\includegraphics[page=2,width=\linewidth]{scatter_results}
		\end{subfigure}
	\end{minipage}
	\caption{Scatter plots of our NDE algorithm compared to other algorithms}
	\label{fig: test_scatter}
\end{figure*}

\begin{figure*}[!t]
	\centering
	\graphicspath{ {figures/nde} }
	\begin{minipage}{1\textwidth}
		\begin{subfigure}[t]{1\textwidth}
			\centering
			\includegraphics[page=1,width=\linewidth]{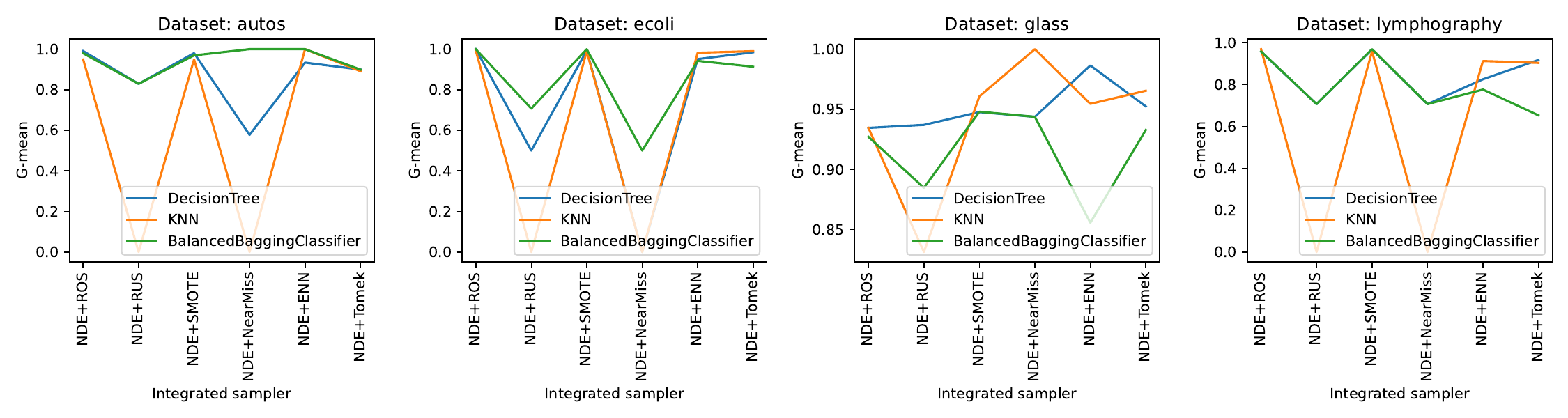}
		\end{subfigure}
	\end{minipage}
	\begin{minipage}{1\textwidth}
		\begin{subfigure}[t]{1\textwidth}
			\centering
			\includegraphics[page=2,width=\linewidth]{sampler_results}
		\end{subfigure}
	\end{minipage}
	\caption{G-mean scores of our NDE algorithm evaluated across three classifiers for various samplers}
	\label{fig: test_sampler}
\end{figure*}

\subsection{Testing procedure}
A given dataset is tested using a resampling method to balance the class distributions. After resampling, the dataset is then partitioned into 80\% training and 20\% testing subsets using cross-validation. The number of cross-validation folds, $n_{\text{splits}}$, is computed as:
\begin{align*}
	n_{\text{splits}} = \min\left(5, \min\left(y_{\text{train}}\right)\right),
\end{align*}
where $\min(y_{train})$ corresponds to the minimum count of samples among the classes in the training data. The classifier is trained using the training portion of the data and subsequently evaluated on the test subset. To assess performance the average G-mean metric is computed across folds which is reported in this study.

\subsection{Results}
We conducted a series of tests to assess the effectiveness of the proposed methods. These tests were divided into two main groups as follows. 

\subsubsection{NDE evaluation}
The first group focused on evaluating the performance of our base algorithm (NDE) using either a single synthetic dataset or eight real-world datasets (indicated by $\clubsuit$ in Table~\ref{tbl: real-data}), along with three classifiers, to visually illustrate the results. This group includes tests to evaluate the effects of varying $k$-neighbors, varying distance metrics, resampled data distributions, and algorithm performance when paired with another sampler for oversampling.	

\begin{table*}[!htp]
	\caption{Error messages encountered during resampling using default parameters across methods}\label{tbl: err-data}
	\footnotesize
	\renewcommand{\arraystretch}{1.2}
	\begin{minipage}{1\textwidth}
		\centering
		\begin{tabular}[t]{L{0.3\linewidth}L{0.25\linewidth}L{0.35\linewidth}}
			\toprule
			\textbf{Failing methods} & \textbf{Dataset} & \textbf{Error messages}\\
			\midrule
			NearMiss, ADASYN, Borderline-SMOTE, SMOTE, SMOTE-ENN, SMOTE-Tomek, SMOTE-CDNN, SVM-SMOTE & autos, ecoli, lymphography, pageblocks, shuttle, yeast (6) & \textit{Expected n\_neighbors <= n\_samples\_fit ...} \\
			SMOTE-ENN, ENN, ECDNN & autos, lymphography, svmguide4, thyroid, vowel, yeast (6) & \textit{The least populated class in y has only 1 member, which is too few ...}\\
			KMeans-SMOTE & autos, balance, ecoli, glass, lymphography, pageblocks, shuttle, thyroid, yeast (9) & \textit{No clusters found with sufficient samples of class ...}\\
			SVM-SMOTE & autos (1) & \textit{All support vectors are considered as noise. SVM-SMOTE is not adapted to your dataset ...}\\
			ADASYN & contraceptive, glass, penbased, svmguide4, vehicle (5) & \textit{No samples will be generated with the provided ratio settings.}\\
			RandomUnder, ENN & ecoli, lymphography, shuttle (3) & \textit{k-fold cross-validation requires at least one train/test split ...}\\
			\bottomrule
		\end{tabular}
	\end{minipage}	
\end{table*}

\begin{enumerate}
	\item \textit{Varying $k$-neighbors}\\
	In this experiment, we evaluated our algorithm on the top 8 datasets with the highest imbalance ratios among the 20 datasets, utilizing three classifiers—DecisionTree, $k$-NN, and Balanced Bagging—while varying the $k$ parameter from 2 to 25. The results, presented in Fig.~\ref{fig: test_k}, demonstrate that our algorithm consistently achieves excellent performance in most cases. Notably, it maintains a G-mean value exceeding 0.90 for $k \geq 5$. These findings underscore the robustness and effectiveness of our approach across varying $k$ values, highlighting its suitability for diverse scenarios.
	\item \textit{Varying distance metric selection}\\
	To assess the robustness of our algorithm, we evaluated its performance across commonly used distance metrics, including \textit{euclidean}, \textit{cityblock}, \textit{minkowski}, \textit{cosine}, and \textit{hamming}. While numerous other metrics could be tested, these were selected due to their prevalence in data processing tasks. For this test, the same 8 datasets and 3 classifiers were utilized, with the default parameters of our algorithm. As shown in Fig.~\ref{fig: test_metrics}, the findings indicate that our algorithm consistently achieves high G-mean values across various datasets and distance metrics. While a slight performance fluctuation is observed with the Hamming distance on the \textit{autos} dataset, the overall stability across metrics underscores the robustness and versatility of our approach in adapting to different metric selections.
	\item \textit{Resampled data distribution}\\
	In this test, we compared the original dataset distribution with the results after applying our algorithm and baseline samplers, including ROS, RUS, SMOTE, ENN, NearMiss, TomekLinks, and ECDNN. Advanced variants such as SVM-SMOTE and SMOTE-Tomek were excluded to focus on the foundational methods. For this test, the synthetic dataset was used, with the default parameters of our algorithm. As shown in Fig.~\ref{fig: test_scatter}, our algorithm best separates data points into isolated clusters, showing its strong performance. While ENN and ECDNN also produce relatively clear distributions, they leave more overlapping points than our method. The other samplers result in significant overlap, producing more noises that can negatively affect classification performance. These results confirm our method's superiority in producing well-separated and cleaner resampled datasets.
	\item \textit{Algorithm in combination with another sampler}\\
	To determine the optimal combination of our algorithm with other samplers, we evaluated the performance of pairing our method with ROS, RUS, SMOTE, ENN, NearMiss, and TomekLinks, excluding ECDNN due to its similar displacement-based approach to ours. The goal was to identify the best sampler for oversampling the data after displacement or removing noisy points. As depicted in Fig.~\ref{fig: test_sampler}, our NDE algorithm pairs effectively with all methods except RUS and NearMiss. As further illustrated in Fig.~\ref{fig: test_scatter}, these two methods tend to discard a significant number of data points, which does not align with our objective of applying oversampling after noisy data points are displaced by our algorithm. In this study, we propose combining NDE with ROS, as it is the simplest approach and delivers excellent results, which we report in the second group of tests.
\end{enumerate}

\subsubsection{NDESO evaluation}
The second group assesses the performance of our NDESO (NDE + ROS) resampling algorithm by comparing it to 14 baseline resampling methods across 20 real-world datasets and nine classifiers, with a focus on G-mean performance and statistical significance. The baseline methods include advanced versions of SMOTE, ADASYN, and SMOTE-CDNN, a hybrid approach combining CDNN with SMOTE oversampling. 		

\begin{table*}[!htp]	
	\raggedright
	\caption{G-mean scores of resampling methods with MLP classifier across datasets}\label{tbl: mlp-results}
	\fontsize{7pt}{6pt}\selectfont
	\begin{minipage}{1\textwidth}
		\setlength{\tabcolsep}{4.5pt}
		\begin{tabular}{@{} lccccccccccccccc @{} }
			\toprule
			\textbf{Dataset} & \textbf{[1]} & \textbf{[2]} & \textbf{[3]} & \textbf{[4]} & \textbf{[5]} & \textbf{[6]} & \textbf{[7]} & \textbf{[8]} & \textbf{[9]} & \textbf{[10]} & \textbf{[11]} & \textbf{[12]} & \textbf{[13]} & \textbf{[14]} & \textbf{[15]}\\
			\midrule
			autos	& \textbf{0.9613}	& 0.6792	& -	& -	& -	& 0.6033	& 0.6875	& 0.9500	& 0.6921	& -	& 0.4274	& 0.6294	& 0.8141	& -	& -\\
			balance	& \textbf{0.9667}	& 0.9160	& 0.8512	& 0.8393	& 0.9351	& 0.9420	& 0.9363	& 0.9505	& 0.9362	& 0.8917	& 0.6434	& 0.6458	& 0.9481	& 0.9343	& -\\
			contraceptive	& \textbf{0.8761}	& 0.5672	& 0.5434	& 0.4942	& -	& 0.5660	& 0.5693	& 0.7466	& 0.5748	& 0.5745	& 0.5433	& 0.6688	& 0.7186	& 0.5785	& -\\
			dermatology	& 0.9962	& 0.9758	& 0.9778	& 0.9500	& 0.9815	& 0.9795	& 0.9794	& 0.9927	& 0.9889	& \textbf{1.0000}	& 0.9709	& 0.9889	& 0.9879	& 0.9943	& 0.9852\\
			ecoli	& \textbf{0.9065}	& 0.8075	& -	& -	& -	& 0.6567	& 0.8538	& 0.8764	& 0.8520	& 0.5702	& 0.3868	& 0.5451	& 0.8841	& -	& -\\
			glass	& \textbf{0.8259}	& 0.7244	& 0.5667	& 0.6667	& -	& 0.7075	& 0.7487	& 0.6707	& 0.7423	& 0.6056	& 0.3744	& 0.4861	& 0.6925	& 0.6462	& 0.7605\\
			hayes-roth	& 0.7380	& 0.6565	& 0.6167	& 0.6633	& 0.6220	& 0.6889	& 0.6741	& \textbf{0.8611}	& 0.6750	& 0.7044	& 0.6521	& 0.8542	& 0.7659	& 0.6333	& 0.6870\\
			lymphography	& 0.9179	& 0.9186	& -	& -	& -	& 0.8990	& 0.9064	& \textbf{0.9521}	& 0.9051	& -	& 0.3868	& -	& 0.9525	& 0.9162	& -\\
			new-thyroid	& \textbf{1.0000}	& 0.9528	& 0.9200	& 0.9600	& 0.9806	& 0.9889	& 0.9500	& 0.9629	& 0.9639	& 0.7211	& 0.7122	& 0.7467	& 0.9690	& 0.9528	& 0.9781\\
			pageblocks	& \textbf{0.9980}	& 0.9777	& -	& -	& 0.9715	& 0.9827	& 0.9837	& 0.9892	& 0.9807	& 0.5714	& 0.3500	& 0.3911	& 0.9884	& 0.9778	& -\\
			penbased	& \textbf{0.9653}	& 0.9458	& 0.9346	& 0.9561	& -	& 0.9563	& 0.9502	& 0.9615	& 0.9562	& 0.9621	& 0.9421	& 0.9482	& 0.9526	& 0.9511	& 0.9506\\
			segment	& \textbf{0.9773}	& 0.9356	& 0.9356	& 0.9340	& 0.9356	& 0.9361	& 0.9367	& 0.9611	& 0.9355	& 0.9603	& 0.9361	& 0.9414	& 0.9494	& 0.9351	& 0.9367\\
			shuttle	& \textbf{0.9918}	& 0.9881	& -	& -	& 0.9811	& 0.9827	& 0.9865	& 0.9950	& 0.9867	& 0.8402	& 0.5632	& 0.5757	& 0.9950	& 0.9817	& -\\
			svmguide2	& 0.8455	& 0.8193	& 0.7120	& 0.6231	& 0.7506	& 0.8437	& 0.8075	& 0.8203	& 0.8364	& 0.8065	& 0.7628	& 0.7915	& 0.8625	& 0.8511	& \textbf{0.8814}\\
			svmguide4	& \textbf{0.9366}	& 0.5106	& 0.5304	& 0.5726	& -	& 0.5718	& 0.4981	& 0.8250	& 0.5248	& -	& 0.4762	& 0.5132	& 0.5733	& 0.4889	& 0.5663\\
			thyroid	& \textbf{0.9806}	& 0.9199	& 0.5889	& 0.6444	& 0.9203	& 0.9355	& 0.9161	& 0.9175	& 0.9148	& -	& 0.4160	& -	& 0.9211	& 0.9011	& -\\
			vehicle	& \textbf{0.9355}	& 0.7637	& 0.7609	& 0.7419	& -	& 0.7608	& 0.7536	& 0.8334	& 0.7705	& 0.8045	& 0.7656	& 0.7585	& 0.7613	& 0.7608	& 0.7763\\
			vowel	& \textbf{0.9828}	& 0.6156	& 0.6114	& 0.5838	& 0.6315	& 0.6140	& 0.6231	& -	& 0.4681	& -	& 0.4598	& 0.4383	& 0.5527	& 0.6019	& 0.6058\\
			wine	& 0.9879	& 0.9758	& 0.9655	& 0.9738	& 0.9889	& 0.9818	& 0.9818	& 0.9889	& 0.9739	& 0.9889	& 0.9793	& \textbf{0.9926}	& 0.9917	& 0.9758	& 0.9828\\
			yeast	& \textbf{0.9349}	& 0.6144	& 0.5750	& 0.3750	& -	& 0.6611	& 0.6362	& 0.6716	& 0.6238	& -	& 0.5356	& 0.7252	& 0.6583	& 0.7000	& -\\
			\midrule
			AVERAGE	& \textbf{0.9362}	& 0.8132	& 0.7393	& 0.7319	& 0.8817	& 0.8129	& 0.8190	& 0.8909	& 0.8151	& 0.7858	& 0.6142	& 0.7023	& 0.8469	& 0.8212	& 0.8283\\
			\bottomrule
		\end{tabular}
		\vspace{-10pt}
		\begin{tablenotes}
			\begin{multicols}{6}
				\item[1][1] NDESO 
				\item[2][2] RandomOver
				\item[3][3] RandomUnder
				\item[4][4] NearMiss
				\item[5][5] ADASYN
				\item[6][6] Borderline-SMOTE
				\item[7][7] SMOTE
				\item[8][8] SMOTE-ENN
				\item[9][9] SMOTE-Tomek
				\item[10][10] ENN
				\item[11][11] TomekLinks
				\item[12][12] ECDNN
				\item[13][13] SMOTE-CDNN
				\item[14][14] SVM-SMOTE
				\item[15][15] KMeans-SMOTE
				\item[16][-] Resampling encountered an error.
			\end{multicols}
		\end{tablenotes}
	\end{minipage}
	\vspace{10pt}
	\begin{minipage}{1\textwidth}
		\includegraphics[width=\linewidth]{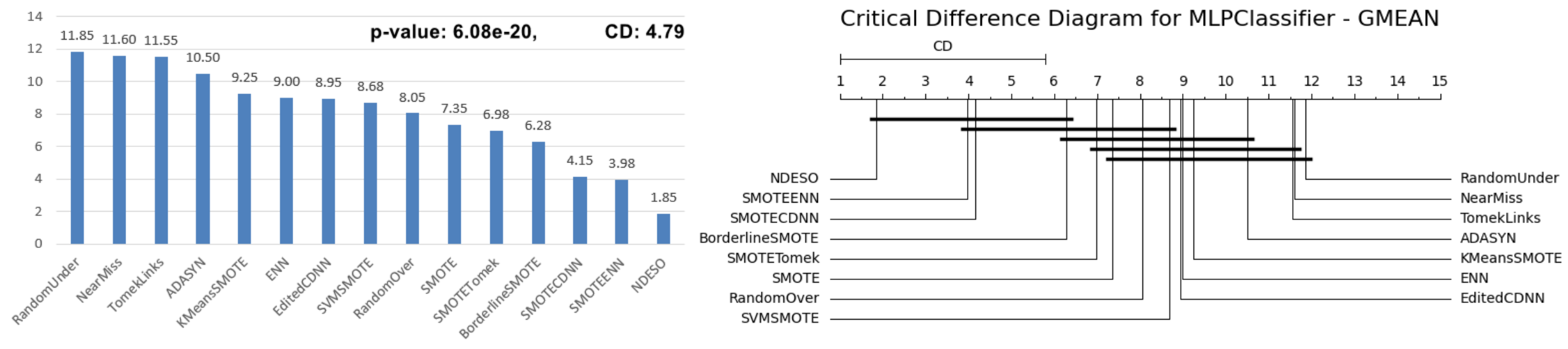}
	\end{minipage}
\end{table*}

\begin{table*}[!htp]	
	\raggedright
	\caption{Execution time (in seconds) of resampling methods with MLP classifier across datasets}\label{tbl: time-results}
	\fontsize{7pt}{6pt}\selectfont
	\begin{minipage}{1\textwidth}
		\setlength{\tabcolsep}{4.5pt}
		\begin{tabular}{@{} lccccccccccccccc @{} }
			\toprule
			\textbf{Dataset} & \textbf{[1]} & \textbf{[2]} & \textbf{[3]} & \textbf{[4]} & \textbf{[5]} & \textbf{[6]} & \textbf{[7]} & \textbf{[8]} & \textbf{[9]} & \textbf{[10]} & \textbf{[11]} & \textbf{[12]} & \textbf{[13]} & \textbf{[14]} & \textbf{[15]}\\
			\midrule
			autos	& 0.0041	& \textbf{0.0024}	& -	& -	& -	& 0.0045	& 0.0062	& 0.0853	& 0.0514	& -	& \textbf{0.0024}	& 0.0049	& 0.0926	& -	& -\\
			balance	& 0.0206	& \textbf{0.0012}	& 0.0023	& 0.0039	& 0.0026	& 0.0024	& 0.0019	& 0.0044	& 0.0039	& 0.0028	& 0.0033	& 0.0157	& 0.0284	& 0.0054	& -\\
			contraceptive	& 0.0964	& 0.0017	& \textbf{0.0015}	& 0.0042	& -	& 0.0077	& 0.0039	& 0.0108	& 0.0093	& 0.0051	& 0.0049	& 0.1101	& 0.1692	& 0.0841	& -\\
			dermatology	& 0.008	& 0.0041	& \textbf{0.0021}	& 0.0037	& 0.0173	& 0.0144	& 0.0051	& 0.0176	& 0.0141	& 0.0053	& 0.0114	& 0.0120	& 0.0276	& 0.0326	& 0.6554\\
			ecoli	& 0.0125	& \textbf{0.0020}	& -	& -	& -	& 0.0123	& 0.0053	& 0.0125	& 0.0088	& 0.0038	& 0.0024	& 0.0082	& 0.0636	& -	& -\\
			glass	& 0.0044	& 0.0018	& \textbf{0.0017}	& 0.0034	& -	& 0.0091	& 0.0045	& 0.0060	& 0.0045	& 0.0033	& 0.0021	& 0.0062	& 0.0124	& 0.0144	& 0.0176\\
			hayes-roth	& 0.0027	& \textbf{0.0014}	& 0.0017	& 0.0024	& 0.0022	& 0.0025	& 0.0018	& 0.0047	& 0.0027	& 0.0028	& 0.0023	& 0.0032	& 0.0051	& 0.0051	& 0.0055\\
			lymphography	& 0.0071	& 0.0024	& -	& -	& -	& 0.0042	& 0.0051	& 0.1013	& 0.0770	& -	& \textbf{0.0021}	& -	& 0.0174	& 0.0840	& -\\
			new-thyroid	& 0.0039	& 0.0020	& 0.0018	& 0.0033	& 0.0029	& 0.0029	& 0.0022	& 0.0037	& 0.0035	& 0.0021	& \textbf{0.0017}	& 0.0085	& 0.0107	& 0.0093	& 0.0078\\
			pageblocks	& 0.0153	& \textbf{0.0016}	& -	& -	& 0.0084	& 0.0550	& 0.0330	& 0.0493	& 0.0350	& 0.0032	& 0.0030	& 0.0141	& 0.2976	& 0.0920	& -\\
			penbased	& 0.0450	& 0.0024	& \textbf{0.0016}	& 0.0938	& -	& 0.0158	& 0.0070	& 0.0866	& 0.0806	& 0.0447	& 0.0749	& 0.0498	& 0.0828	& 0.1453	& 0.8824\\
			segment	& 0.2421	& \textbf{0.0023}	& 0.0029	& 0.0533	& 0.0023	& 0.0033	& 0.0021	& 0.0273	& 0.0061	& 0.0151	& 0.0072	& 0.2694	& 0.2516	& 0.0030	& 0.0020\\
			shuttle	& 0.2376	& \textbf{0.0020}	& -	& -	& 0.0104	& 0.0128	& 0.0071	& 0.0734	& 0.0680	& 0.0170	& 0.0210	& 0.1977	& 3.0294	& 0.0651	& -\\
			svmguide2	& 0.0086	& 0.0039	& \textbf{0.0025}	& 0.0044	& 0.0864	& 0.0754	& 0.0749	& 0.0849	& 0.0799	& 0.0730	& 0.0743	& 0.0099	& 0.0373	& 0.0914	& 0.1012\\
			svmguide4	& 0.0069	& 0.0025	& \textbf{0.0014}	& 0.0034	& -	& 0.0053	& 0.0041	& 0.0057	& 0.0052	& -	& 0.0031	& 0.0099	& 0.0136	& 0.0283	& 0.1271\\
			thyroid	& 0.0206	& 0.0029	& \textbf{0.0017}	& 0.0914	& 0.0732	& 0.0862	& 0.0048	& 0.0077	& 0.0076	& -	& 0.0099	& -	& 0.2731	& 0.1068	& -\\
			vehicle	& 0.0337	& 0.0019	& \textbf{0.0017}	& 0.0060	& -	& 0.0060	& 0.0822	& 0.0873	& 0.0805	& 0.0766	& 0.0615	& 0.0332	& 0.1361	& 0.1210	& 0.3299\\
			vowel	& 0.0138	& 0.0020	& \textbf{0.0015}	& 0.0047	& 0.0014	& 0.0014	& 0.0016	& -	& 0.0054	& -	& 0.0028	& 0.0293	& 0.0176	& 0.0016	& 0.0016\\
			wine	& 0.0047	& 0.0017	& \textbf{0.0015}	& 0.0027	& 0.0040	& 0.0040	& 0.0040	& 0.0064	& 0.0039	& 0.0024	& 0.0038	& 0.0072	& 0.0072	& 0.0062	& 0.0138\\
			yeast	& 0.1244	& 0.0019	& \textbf{0.0015}	& 0.0719	& -	& 0.0572	& 0.0162	& 0.0882	& 0.0303	& -	& 0.0075	& 0.0905	& 1.0080	& 0.4161	& -\\
			\midrule
			AVERAGE	& 0.0456	& 0.0022	& \textbf{0.0018}	& 0.0235	& 0.0192	& 0.0191	& 0.0137	& 0.0402	& 0.0289	& 0.0184	& 0.0151	& 0.0489	& 0.2791	& 0.0729	& 0.1950\\
			\bottomrule
		\end{tabular}
		\vspace{-10pt}
		\begin{tablenotes}
			\begin{multicols}{6}
				\item[1][1] NDESO 
				\item[2][2] RandomOver
				\item[3][3] RandomUnder
				\item[4][4] NearMiss
				\item[5][5] ADASYN
				\item[6][6] Borderline-SMOTE
				\item[7][7] SMOTE
				\item[8][8] SMOTE-ENN
				\item[9][9] SMOTE-Tomek
				\item[10][10] ENN
				\item[11][11] TomekLinks
				\item[12][12] ECDNN
				\item[13][13] SMOTE-CDNN
				\item[14][14] SVM-SMOTE
				\item[15][15] KMeans-SMOTE
				\item[16][-] Resampling encountered an error.
			\end{multicols}
		\end{tablenotes}
	\end{minipage}
\end{table*}

\begin{table*}[!htp]
	\raggedright
	\caption{G-mean scores of resampling methods across various classifiers}\label{tbl: classifier-results}		
	\begin{minipage}{1\textwidth}
		\fontsize{7pt}{6pt}\selectfont
		\setlength{\tabcolsep}{3.5pt}
		\begin{tabular}{@{} lccccccccccccccc @{} }
			\toprule
			\textbf{Classifiers} & \textbf{[1]} & \textbf{[2]} & \textbf{[3]} & \textbf{[4]} & \textbf{[5]} & \textbf{[6]} & \textbf{[7]} & \textbf{[8]} & \textbf{[9]} & \textbf{[10]} & \textbf{[11]} & \textbf{[12]} & \textbf{[13]} & \textbf{[14]} & \textbf{[15]}\\
			\midrule
			SVC	& \textbf{0.9389}	& 0.8320	& 0.7506	& 0.7455	& 0.9072	& 0.8256	& 0.8420	& 0.8962	& 0.8385	& 0.8152	& 0.6468	& 0.7381	& 0.8585	& 0.8418	& 0.8652 \\
			$k$-NN	& \textbf{0.9324}	& 0.8200	& 0.6781	& 0.6759	& 0.8675	& 0.8086	& 0.8285	& 0.9000	& 0.8203	& 0.7884	& 0.6258	& 0.6939	& 0.8543	& 0.8267	& 0.8221 \\
			Decision Tree	& \textbf{0.9497}	& 0.8811	& 0.7192	& 0.6812	& 0.9035	& 0.8452	& 0.8577	& 0.9171	& 0.8603	& 0.8571	& 0.7189	& 0.7853	& 0.8929	& 0.8528	& 0.8581 \\
			Random Forest	& \textbf{0.9706}	& 0.9204	& 0.7914	& 0.7560	& 0.9484	& 0.8965	& 0.9119	& 0.9666	& 0.9087	& 0.8761	& 0.7495	& 0.8229	& 0.9332	& 0.9058	& 0.9160 \\
			MLP	& \textbf{0.9362}	& 0.8132	& 0.7393	& 0.7319	& 0.8817	& 0.8129	& 0.8190	& 0.8909	& 0.8151	& 0.7858	& 0.6142	& 0.7023	& 0.8469	& 0.8212	& 0.8283 \\
			Easy Ensemble	& \textbf{0.8183}	& 0.6942	& 0.6333	& 0.6134	& 0.7794	& 0.6975	& 0.7110	& 0.8028	& 0.7283	& 0.7787	& 0.6688	& 0.7189	& 0.7638	& 0.7483	& 0.7018 \\
			RUS Boost	& 0.6690	& 0.5509	& 0.5744	& 0.5498	& 0.6658	& 0.6116	& 0.5746	& 0.7269	& 0.6104	& \textbf{0.7625}	& 0.6340	& 0.6658	& 0.6961	& 0.6618	& 0.5648 \\
			Balanced Bagging	& \textbf{0.9592}	& 0.8954	& 0.7546	& 0.7327	& 0.9201	& 0.8590	& 0.8886	& 0.9238	& 0.8885	& 0.8561	& 0.7416	& 0.8091	& 0.9084	& 0.8770	& 0.8860 \\
			Balanced Random Forest	& \textbf{0.9714}	& 0.9191	& 0.7987	& 0.7545	& 0.9479	& 0.8746	& 0.9109	& 0.9376	& 0.9079	& 0.8998	& 0.7737	& 0.822	& 0.9339	& 0.9033	& 0.9159 \\
			\midrule
			AVERAGE	& \textbf{0.9051}	& 0.8140	& 0.7155	& 0.6934	& 0.8691	& 0.8035	& 0.8160	& 0.8847	& 0.8198	& 0.8244	& 0.6859	& 0.7509	& 0.8542	& 0.8265	& 0.8176 \\							
			\bottomrule
		\end{tabular}						
	\end{minipage}		
	\par\vspace{10pt}
	\small Mean rank results of resampling methods across various classifiers
	\begin{minipage}{1\textwidth}
		\fontsize{7pt}{6pt}\selectfont
		\setlength{\tabcolsep}{6.5pt}
		\begin{tabular}{@{} lccccccccccccccc @{} }
			\toprule
			\textbf{Classifiers} & \textbf{[1]} & \textbf{[2]} & \textbf{[3]} & \textbf{[4]} & \textbf{[5]} & \textbf{[6]} & \textbf{[7]} & \textbf{[8]} & \textbf{[9]} & \textbf{[10]} & \textbf{[11]} & \textbf{[12]} & \textbf{[13]} & \textbf{[14]} & \textbf{[15]}\\
			\midrule
			SVC	& \textbf{2.88}	& 7.75	& 12.10	& 11.50	& 10.30	& 7.58	& 7.00	& 3.28	& 6.88	& 8.95	& 11.30	& 8.18	& 4.30	& 8.03	& 10.00 \\
			$k$-NN	& \textbf{2.10}	& 7.55	& 11.85	& 11.45	& 10.68	& 7.03	& 7.18	& 3.63	& 7.58	& 8.53	& 11.45	& 8.88	& 3.90	& 8.05	& 10.18 \\
			Decision Tree	& \textbf{2.35}	& 4.73	& 11.38	& 12.03	& 11.03	& 7.73	& 7.68	& 4.88	& 7.43	& 8.38	& 10.40	& 9.33	& 4.15	& 8.45	& 10.10 \\
			Random Forest	& \textbf{2.23}	& 5.75	& 12.20	& 12.65	& 10.03	& 7.60	& 6.90	& 3.73	& 7.00	& 8.20	& 11.30	& 9.03	& 4.53	& 8.40	& 10.48 \\
			MLP	& \textbf{1.85}	& 8.05	& 11.85	& 11.60	& 10.50	& 6.28	& 7.35	& 3.98	& 6.98	& 9.00	& 11.55	& 8.95	& 4.15	& 8.68	& 9.25 \\
			Easy Ensemble	& \textbf{3.80}	& 8.58	& 11.50	& 11.18	& 10.70	& 7.70	& 8.15	& 5.53	& 6.65	& 8.73	& 8.85	& 6.63	& 4.00	& 7.85	& 10.18 \\
			RUS Boost	& 6.18	& 9.98	& 10.78	& 11.28	& 10.83	& 7.50	& 8.78	& 5.33	& 7.08	& 6.55	& 6.85	& 5.73	& \textbf{4.10}	& 7.90	& 11.18 \\
			Balanced Bagging	& \textbf{2.40}	& 4.98	& 12.08	& 12.00	& 10.93	& 7.30	& 6.83	& 4.88	& 6.30	& 9.73	& 10.45	& 8.83	& 4.70	& 8.88	& 9.75 \\
			Balanced Random Forest	& \textbf{2.18}	& 5.73	& 11.35	& 12.00	& 11.23	& 7.95	& 7.55	& 3.60	& 7.58	& 8.95	& 10.95	& 8.58	& 3.98	& 7.95	& 10.45 \\
			\midrule
			AVERAGE	& \textbf{2.88}	& 7.01	& 11.68	& 11.74	& 10.69	& 7.41	& 7.49	& 4.31	& 7.05	& 8.56	& 10.34	& 8.23	& 4.20	& 8.24	& 10.17 \\				
			\bottomrule
		\end{tabular}
		\vspace{-10pt}
		\begin{tablenotes}
			\begin{multicols}{6}
				\item[1][1] NDESO 
				\item[2][2] RandomOver
				\item[3][3] RandomUnder
				\item[4][4] NearMiss
				\item[5][5] ADASYN
				\item[6][6] Borderline-SMOTE
				\item[7][7] SMOTE
				\item[8][8] SMOTE-ENN
				\item[9][9] SMOTE-Tomek
				\item[10][10] ENN
				\item[11][11] TomekLinks
				\item[12][12] ECDNN
				\item[13][13] SMOTE-CDNN
				\item[14][14] SVM-SMOTE
				\item[15][15] KMeans-SMOTE
			\end{multicols}
		\end{tablenotes}
	\end{minipage}
\end{table*}

\begin{table*}[!htp]
	\caption{The effect of varying $k$-neighbors on our NDESO algorithm tested across datasets and classifiers}\label{tbl: ndeso-k}
	\begin{minipage}{1\textwidth}
		\fontsize{7pt}{6pt}\selectfont
		\setlength{\tabcolsep}{6.5pt}
		\begin{tabular}{@{} lcccccc @{} }
			\toprule
			\textbf{Classifier} & \textbf{NDESO(k=2)} & \textbf{NDESO(k=5)} & \textbf{NDESO(k=11)} & \textbf{NDESO(k=15)}  & \textbf{NDESO(k=21)} & \textbf{NDESO(k=25)} \\
			\midrule
			SVC	& 0.9032	& 0.9389	& \textbf{0.9398}	& 0.9253	& 0.9231	& 0.9204 \\
			$k$-NN	& 0.8818	& 0.9324	& \textbf{0.9375}	& 0.9294	& 0.9279	& 0.9247 \\
			Decision Tree	& 0.9166	& 0.9487	& \textbf{0.9518}	& 0.9406	& 0.9429	& 0.9473 \\
			Random Forest	& 0.9511	& \textbf{0.9706}	& 0.9694	& 0.9589	& 0.9628	& 0.9626 \\
			MLP	& 0.8888	& \textbf{0.9332}	& 0.9283	& 0.9136	& 0.9120	& 0.9111 \\
			Easy Ensemble	& 0.7710	& \textbf{0.8157}	& 0.8080	& 0.7944	& 0.7811	& 0.7906 \\
			RUS Boost	& 0.6046	& \textbf{0.6645}	& 0.6541	& 0.6635	& 0.6466	& 0.6527 \\
			Balanced Bagging	& 0.9313	& \textbf{0.9576}	& 0.9568	& 0.9471	& 0.9493	& 0.9530 \\
			Balanced Random Forest	& 0.9490	& \textbf{0.9711}	& 0.9707	& 0.9585	& 0.9644	& 0.9632 \\
			\midrule
			AVERAGE	& 0.8664	& \textbf{0.9036}	& 0.9018	& 0.8924	& 0.8900	& 0.8917 \\
			\bottomrule
		\end{tabular}
	\end{minipage}	
\end{table*}

Initially, we compared our method with all other resampling techniques, using their default parameters for a fair evaluation. However, while our method is flawless for any $k$-neighbor selection, some existing algorithms fail due to the sparse distribution of minority class data points in many datasets. These failures were primarily caused by the inability of the algorithms to handle situations where the number of neighbors is lower than the required minimum. A summary of such results is shown in Table~\ref{tbl: err-data}. As observed, KMeans-SMOTE is the most unstable when using default parameters, failing to execute on nine datasets. It is followed by NearMiss, ADASYN, Borderline-SMOTE, SMOTE, SMOTE-ENN, SMOTE-Tomek, SMOTE-CDNN, and SVM-SMOTE, all of which expect the minority class to have at least five neighbors. SMOTE-ENN, ENN, and ECDNN failed on several datasets where a class had only a single member. Similarly, RandomUnder and ENN also encountered failures during internal cross-validation when applied to small subsets of datasets. The only datasets where these methods successfully executed with default parameters are \textit{dermatology}, \textit{hayes-roth}, \textit{new-thyroid}, \textit{segment}, \textit{svmguide2}, and \textit{wine}.

To continue testing, we adjusted the $k$-neighbors parameter for methods like ADASYN, Borderline-SMOTE, SMOTE, SMOTE-ENN, SMOTE-Tomek, ENN, ECDNN, SVM-SMOTE, and KMeans-SMOTE. If resampling failed on the initial attempt, the number of neighbors was progressively reduced until the sampler succeeded or reached the lowest possible value of $k$. However, some samplers, such as ADASYN, SVM-SMOTE, and KMeans-SMOTE, were still unable to process some datasets even after reducing $k$ to its lowest value.

The final evaluation results, summarized in Table~\ref{tbl: mlp-results}, highlight the performance of our approach. While the resampling was tested across various classifiers, yielding consistent overall superiority (as further detailed in subsequent analyses), this table explicitly showcases the best performance observed when using the MLP Classifier. As indicated in the table, our NDESO algorithm achieved the highest average G-mean of 0.9362, surpassing other algorithms in most cases. It exhibited only slightly lower performance on 5 out of 20 datasets, namely \textit{dermatology}, \textit{hayes-roth}, \textit{lymphography}, \textit{svmguide2}, and \textit{wine}. Furthermore, our proposed method demonstrates significant improvements, mainly when applied to the \textit{autos}, \textit{contraceptive}, \textit{svmguide4}, \textit{vowel}, and \textit{yeast} datasets. On these datasets, other methods yield average results between 0.40 and 0.70, whereas our approach consistently achieves averages of 0.90 or higher. This indicates that our algorithm is more effective in reducing data overlap and performs better resampling to generate a more balanced and representative dataset for each class.

The Friedman and Nemenyi post-hoc test results, shown at the bottom of the table, provide statistical validation for the observed improvement. With a p-value of 6.08e-20, the results strongly reject the null hypothesis, confirming that the differences are highly unlikely to have occurred by chance. The differences between methods are significant if the difference in the mean rank is greater than the Critical Difference (CD) = 4.796 of the Nemenyi test. From this, it can be concluded that no significant differences exist among the following groups:
\begin{itemize}
	\item RandomUnder, NearMiss, TomekLinks, ADASYN, KMeans-SMOTE, ENN, ECDNN, SVM-SMOTE, RandomOver, and SMOTE
	\item NearMiss, TomekLinks, ADASYN, KMeans-SMOTE, ENN, ECDNN, SVM-SMOTE, RandomOver, SMOTE, and SMOTE-Tomek
	\item ADASYN, KMeans-SMOTE, ENN, ECDNN, SVM-SMOTE, RandomOver, SMOTE, SMOTE-Tomek, and Borderline-SMOTE
	\item SVM-SMOTE, RandomOver, SMOTE, SMOTE-Tomek, Borderline-SMOTE, SMOTE-CDNN, and SMOTE-ENN
	\item Borderline-SMOTE, SMOTE-CDNN, SMOTE-ENN, and NDESO
\end{itemize} 
All other differences are statistically significant. This means that while our method is comparable to Borderline-SMOTE, SMOTE-CDNN, and SMOTE-ENN, with its lowest average mean rank of 1.85 compared to other resamplers, it signifies better performance as evidenced by clear separation with a substantial margin apart from them in the CD diagram. In addition, our algorithm achieved an average G-mean score of 0.9362, higher than Borderline-SMOTE's average of 0.8129, SMOTE-ENN's average of 0.8909 (with one failure on the \textit{vowel} dataset) and SMOTE-CDNN's average of 0.8469, emphasizes its excellence.

The performance trend is consistently achieved by our algorithm, as shown in the testing across all datasets and classifiers, summarized in Table~\ref{tbl: classifier-results}. Our algorithm scored the highest average G-mean of 0.9051 across all classifiers. With an average mean rank of 2.88, the lowest among all methods, it outperforms other algorithms.

\begin{figure*}[!htp]
	\centering
	\graphicspath{ {figures/ndeso} }
	\begin{minipage}[b]{0.45\textwidth}
		\includegraphics[page=1,width=\textwidth]{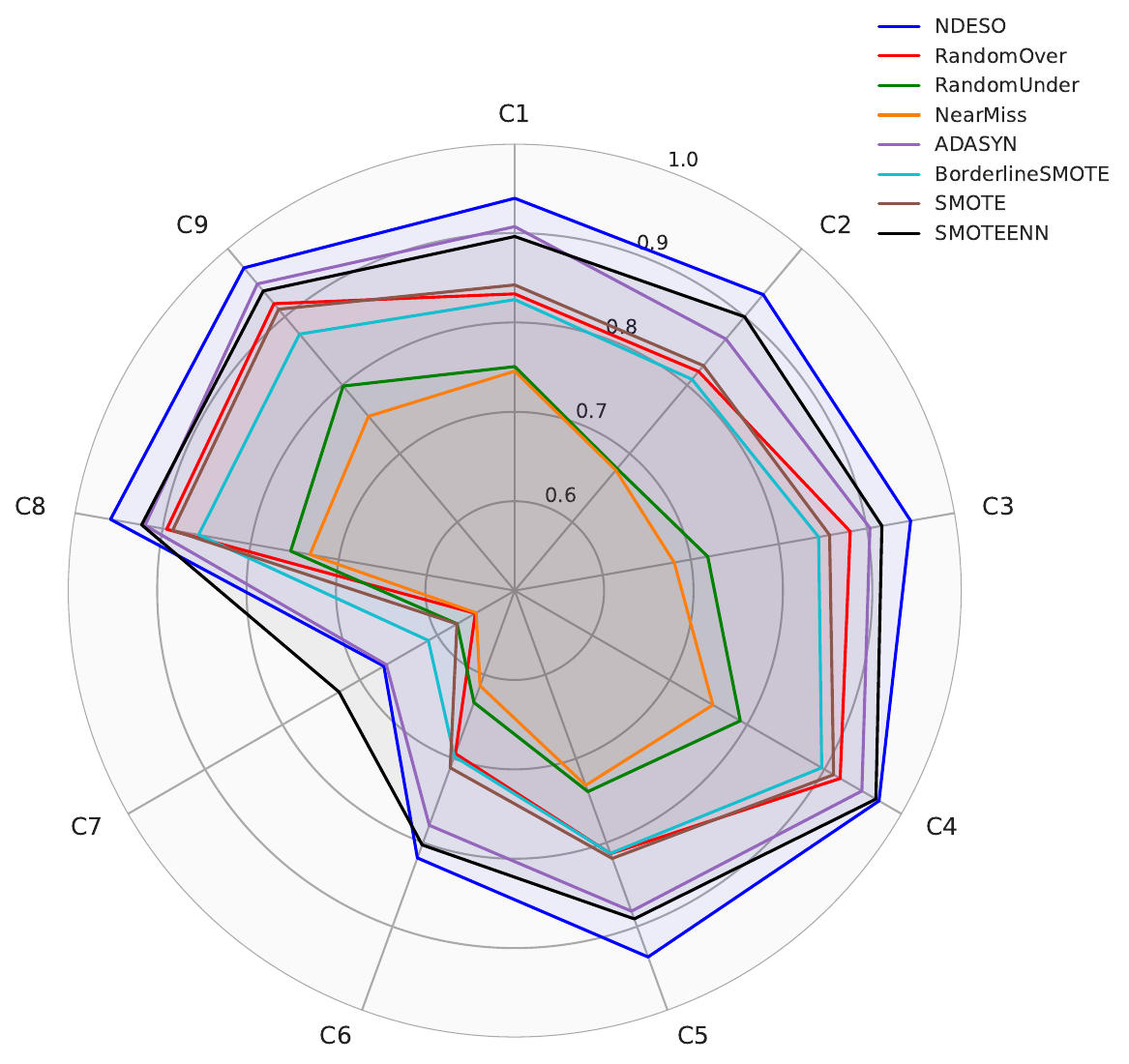}
	\end{minipage}%
	\hspace{0.01\textwidth}
	\begin{minipage}[b]{0.45\textwidth}
		\includegraphics[page=2,width=\textwidth]{radar}
	\end{minipage}
	\hspace{0.01\textwidth}
	\begin{minipage}[b]{0.45\textwidth}
		\includegraphics[page=3,width=\textwidth]{radar}
	\end{minipage}
	\caption{Comparison of average G-mean scores of resampling methods across classifiers and datasets}
	\label{fig: test_radar}
\end{figure*}	

\begin{figure*}[!htp]
	\centering
	\graphicspath{ {figures/ndeso} }
	\begin{minipage}{1\textwidth}
		\begin{subfigure}[t]{1\textwidth}
			\centering
			\includegraphics[page=1,width=\linewidth]{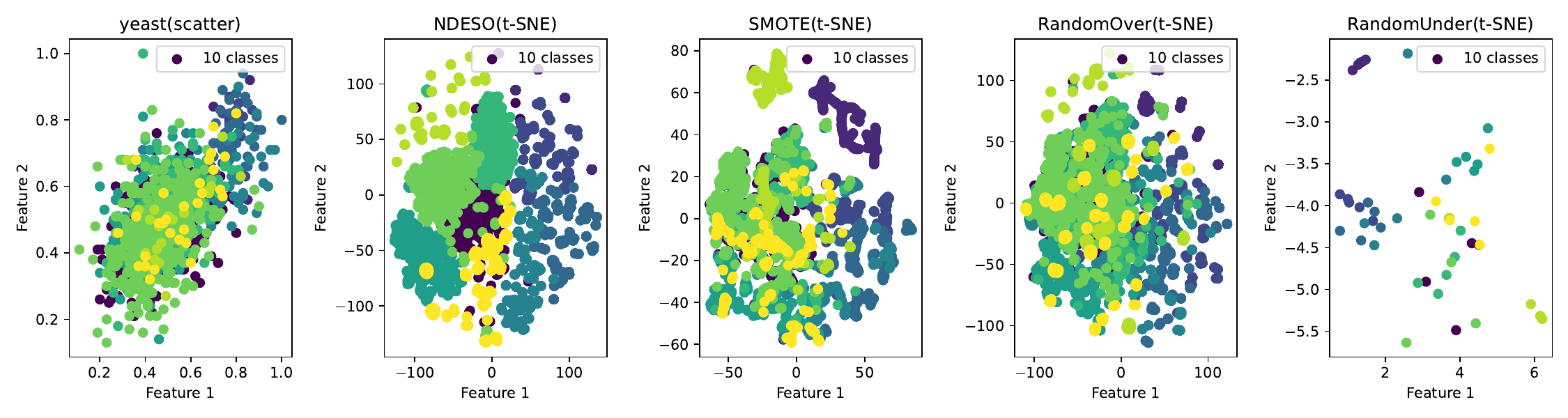}
		\end{subfigure}
	\end{minipage}
	\begin{minipage}{1\textwidth}
		\begin{subfigure}[t]{1\textwidth}
			\centering
			\includegraphics[page=2,width=\linewidth]{scatter}
		\end{subfigure}
	\end{minipage}
	\begin{minipage}{1\textwidth}
		\begin{subfigure}[t]{1\textwidth}
			\centering
			\includegraphics[page=3,width=\linewidth]{scatter}
		\end{subfigure}
	\end{minipage}
	\caption{Scatter plots of resampling methods applied to the \textit{yeast} dataset}
	\label{fig: test_scatter2}
\end{figure*}	

\begin{figure*}[!htp]
	\centering
	\graphicspath{ {figures/cm} }
	\begin{minipage}{1\textwidth}
		\begin{subfigure}[t]{0.25\textwidth}
			\centering
			\includegraphics[page=1,width=\linewidth]{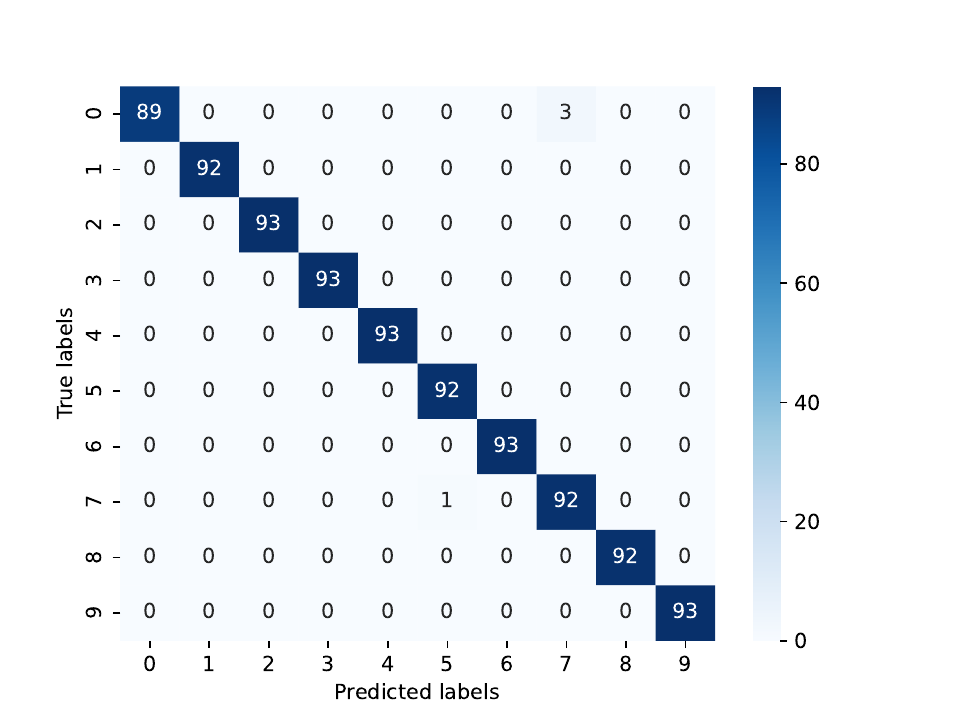}
			\caption*{NDESO}
		\end{subfigure}%
		\begin{subfigure}[t]{0.25\textwidth}
			\centering
			\includegraphics[page=1,width=\linewidth]{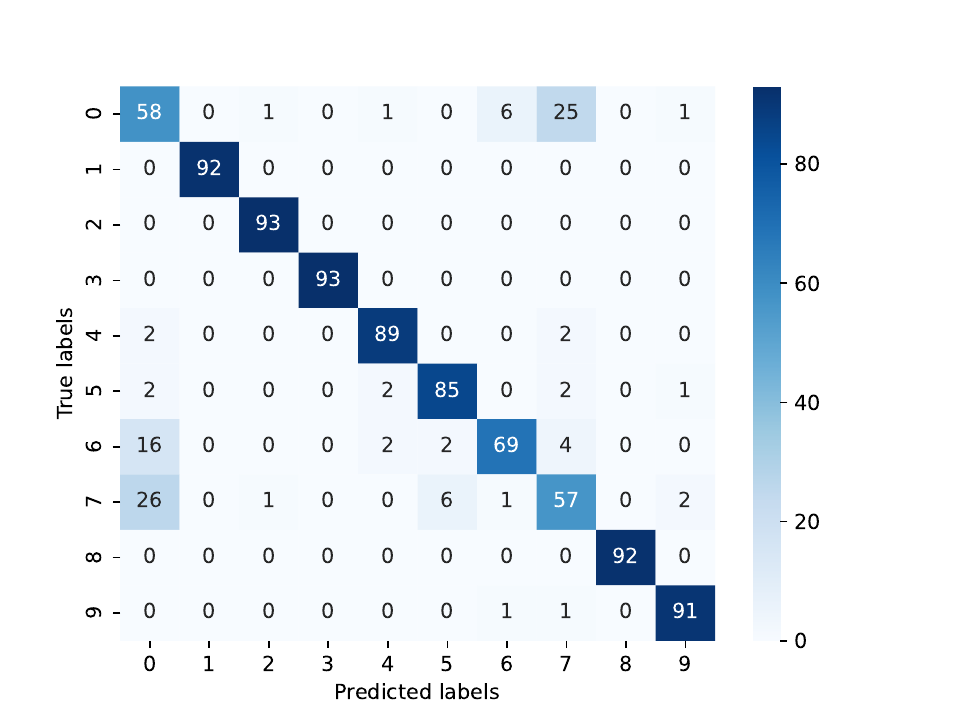}
			\caption*{SMOTE}
		\end{subfigure}%
		\begin{subfigure}[t]{0.25\textwidth}
			\centering
			\includegraphics[page=1,width=\linewidth]{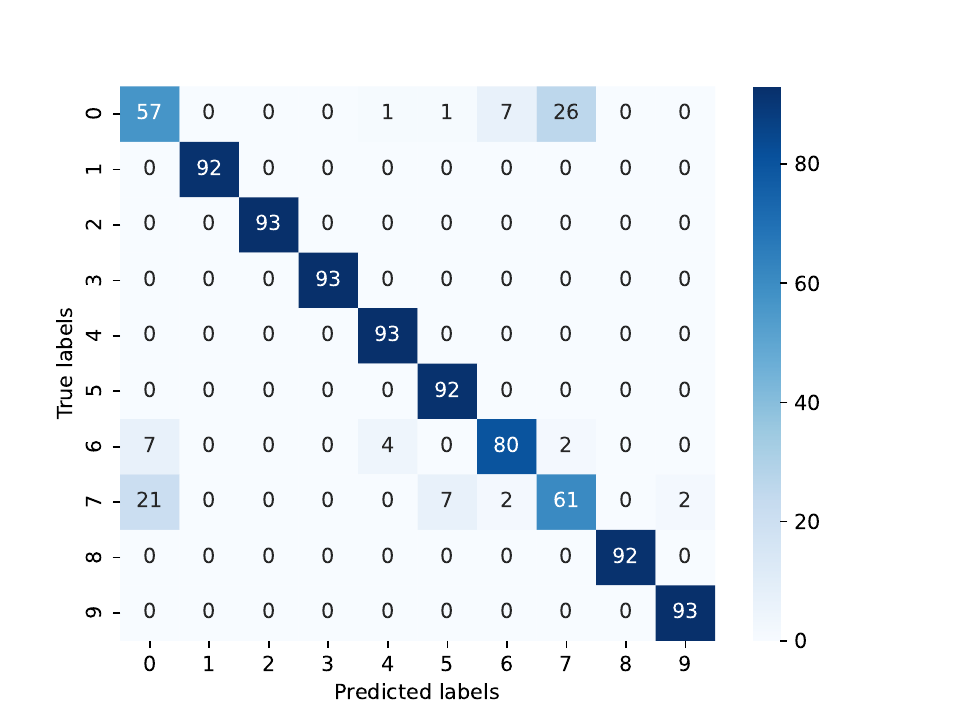}
			\caption*{RandomOver}
		\end{subfigure}%
		\begin{subfigure}[t]{0.25\textwidth}
			\centering
			\includegraphics[page=1,width=\linewidth]{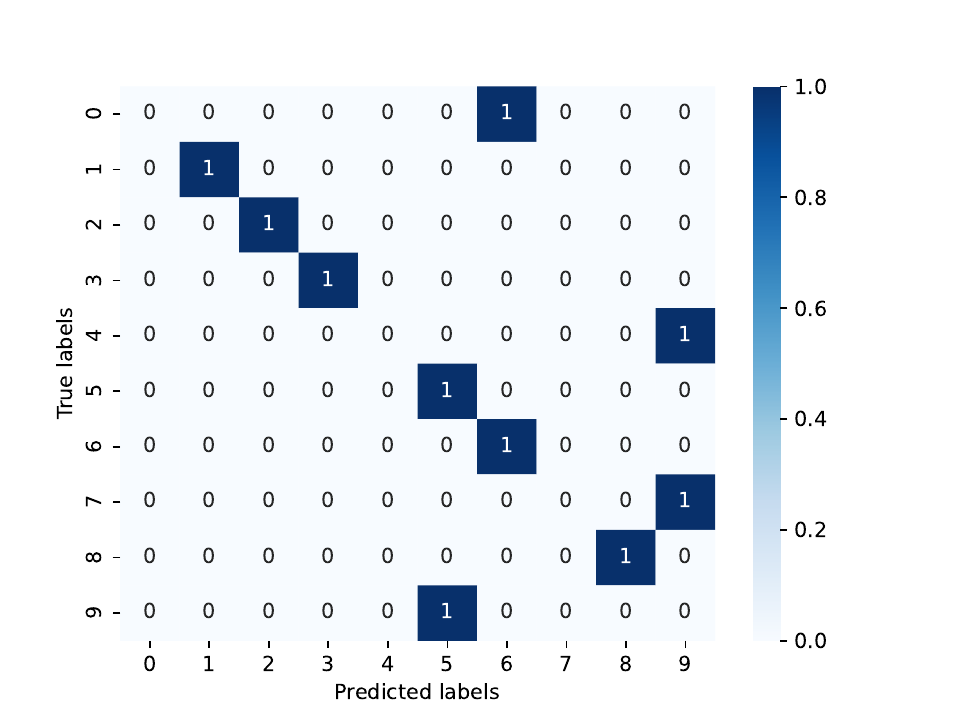}
			\caption*{RandomUnder}
		\end{subfigure}
	\end{minipage}
	\begin{minipage}{1\textwidth}
		\begin{subfigure}[t]{0.25\textwidth}
			\centering
			\includegraphics[page=1,width=\linewidth]{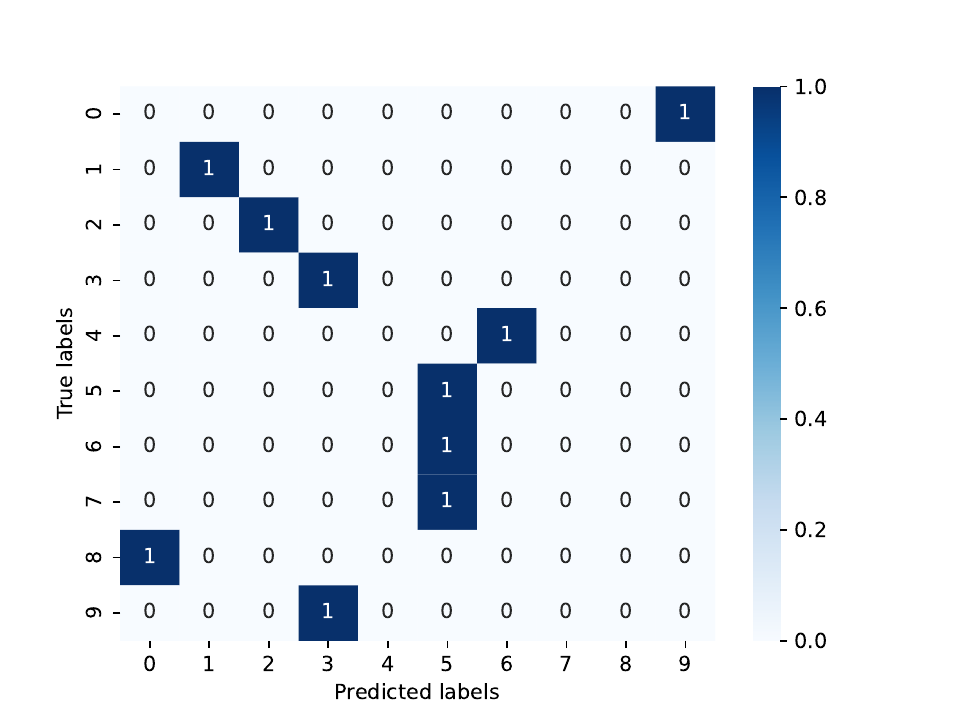}
			\caption*{NearMiss}
		\end{subfigure}%
		\begin{subfigure}[t]{0.25\textwidth}
			\centering
			\includegraphics[page=1,width=\linewidth]{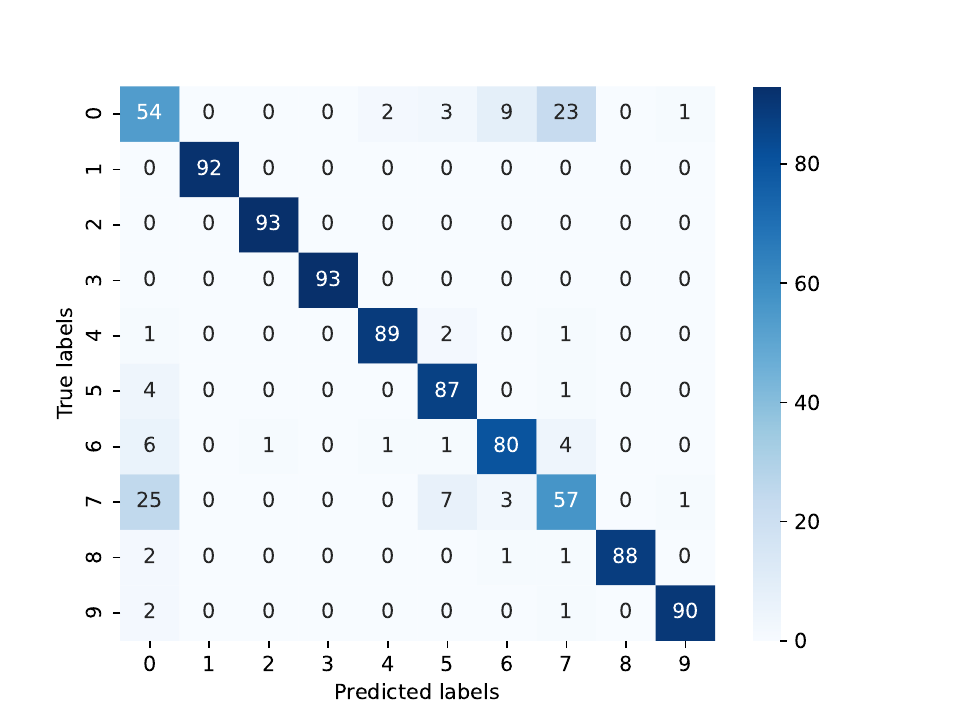}
			\caption*{Borderline-SMOTE}
		\end{subfigure}%
		\begin{subfigure}[t]{0.25\textwidth}
			\centering
			\includegraphics[page=1,width=\linewidth]{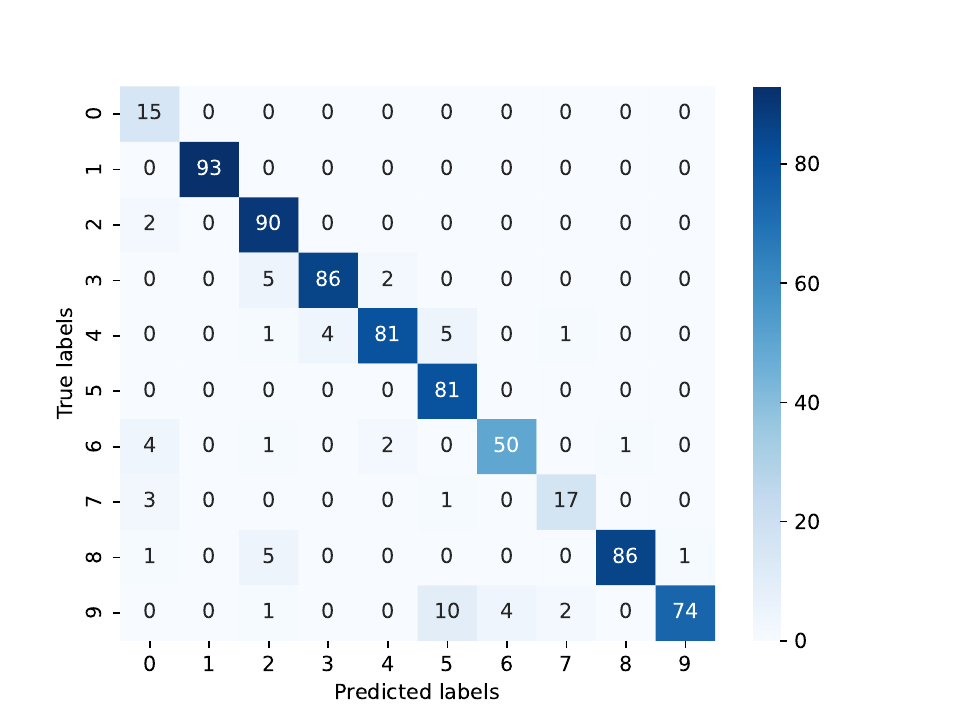}
			\caption*{SMOTE-ENN}
		\end{subfigure}%
		\begin{subfigure}[t]{0.25\textwidth}
			\centering
			\includegraphics[page=1,width=\linewidth]{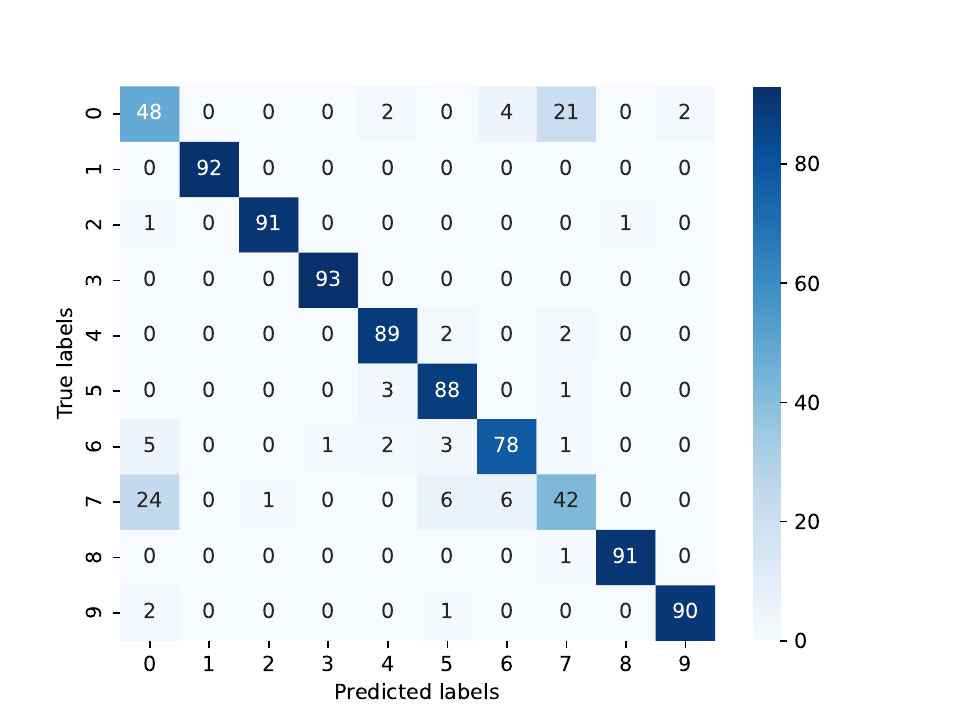}
			\caption*{SMOTE-Tomek}
		\end{subfigure}
	\end{minipage}
	\begin{minipage}{1\textwidth}
		\begin{subfigure}[t]{0.25\textwidth}
			\centering
			\includegraphics[page=1,width=\linewidth]{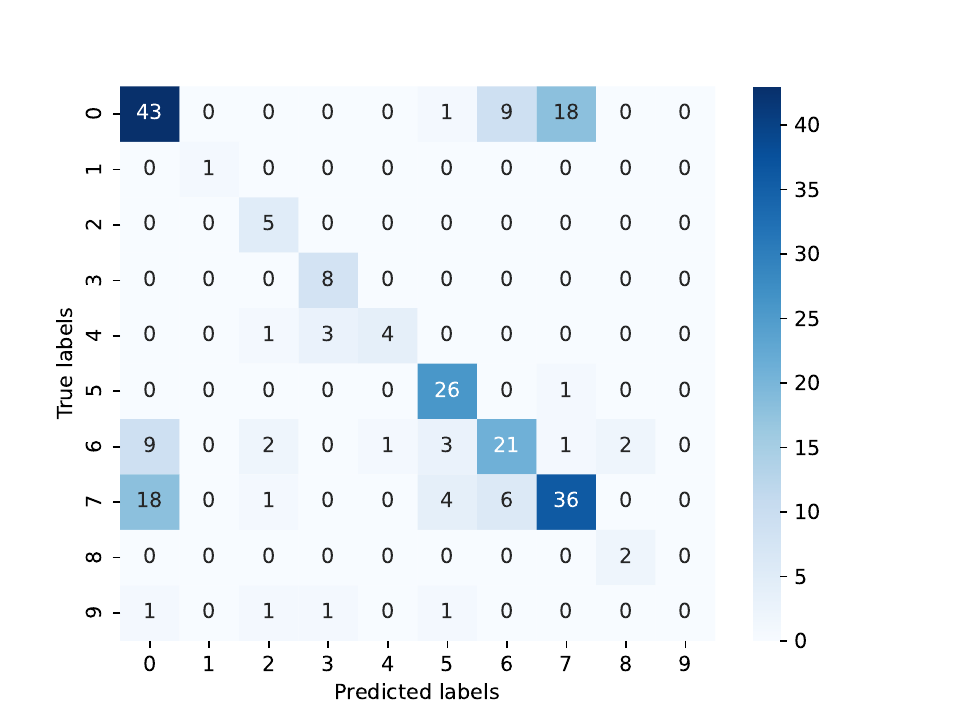}
			\caption*{TomekLinks}
		\end{subfigure}%
		\begin{subfigure}[t]{0.25\textwidth}
			\centering
			\includegraphics[page=1,width=\linewidth]{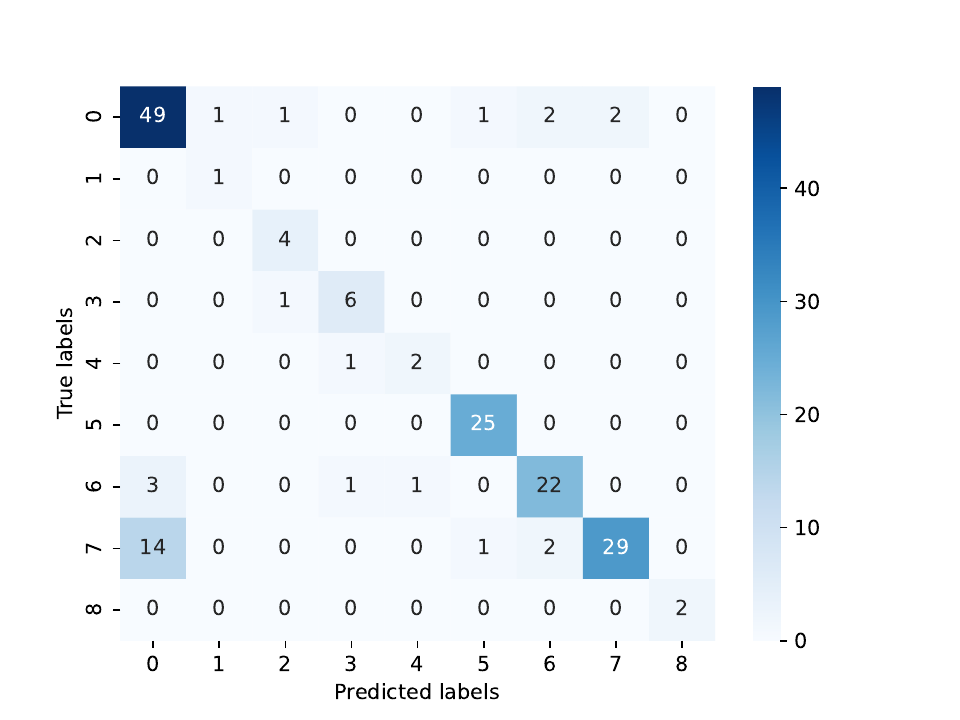}
			\caption*{ECDNN}
		\end{subfigure}%
		\begin{subfigure}[t]{0.25\textwidth}
			\centering
			\includegraphics[page=1,width=\linewidth]{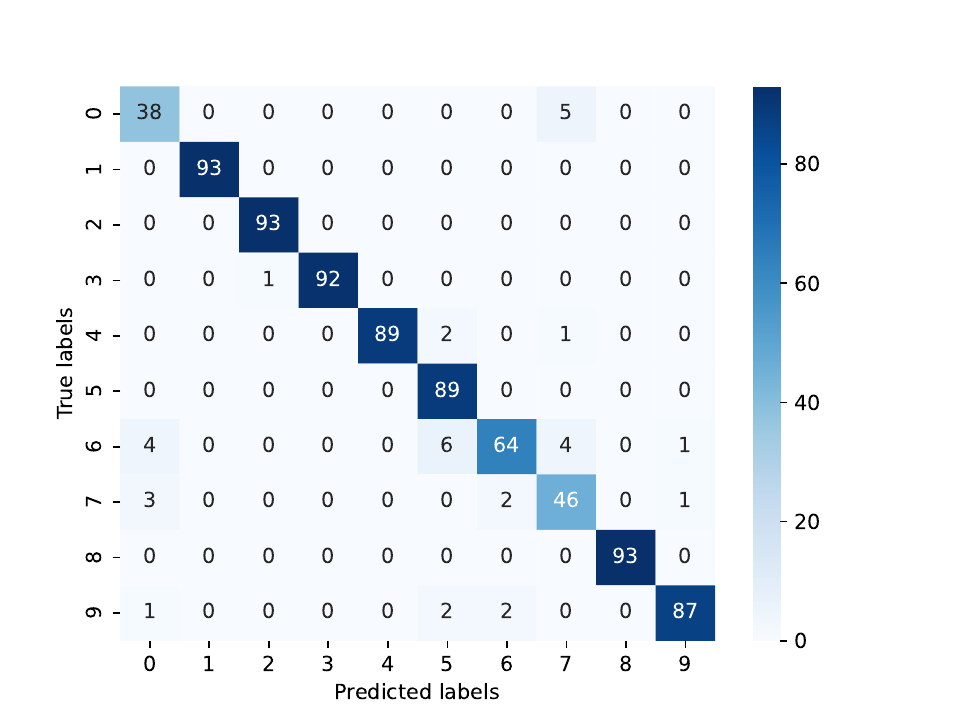}
			\caption*{SMOTE-CDNN}
		\end{subfigure}%
		\begin{subfigure}[t]{0.25\textwidth}
			\centering
			\includegraphics[page=1,width=\linewidth]{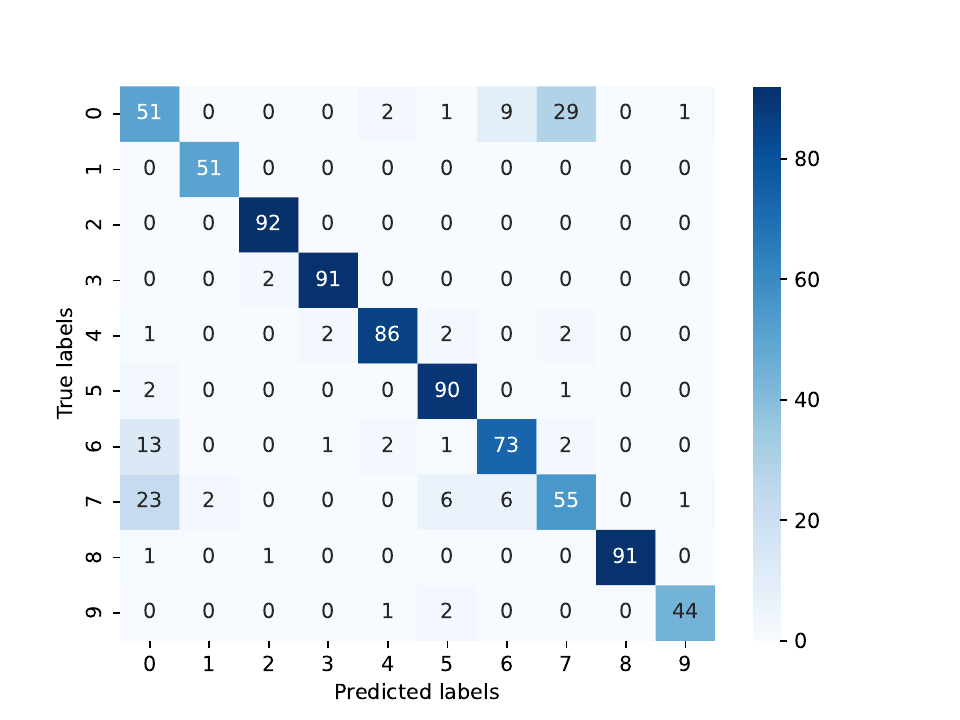}
			\caption*{SVM-SMOTE}
		\end{subfigure}
	\end{minipage}
	\caption{Confusion matrices for resampling methods evaluated using G-mean scores on \textit{yeast} dataset}
	\label{fig: test_cm}
\end{figure*}

We also conducted tests to demonstrate the robustness of our algorithm across different selections of $k$-neighbors = [2, 5, 11, 21, 25], as shown in Table~\ref{tbl: ndeso-k}. The results indicate that our algorithm consistently performs well across different values of $k$, with $k=5$ showing the best performance with an average G-mean of 0.9036, corresponding to our algorithm's default parameter. This suggests that the approach remains effective even with varying $k$ values across datasets and classifiers.

The radar plot presented in Fig.~\ref{fig: test_radar} demonstrates the consistently superior performance of our NDESO algorithm compared to other resampling methods across various datasets and classifiers. As illustrated in the plot, the line representing our algorithm is consistently positioned at the outermost edge in most classifications, indicating that it outperforms all other methods in terms of G-mean across the tested scenarios. This visually reinforces the effectiveness and robustness of NDESO, providing a clear indication of its superior classification results. However, using the RUS Boost classifier appears to be less effective for sparse imbalanced data, especially for imbalanced ratio with majority classes more than 54-56\% \citep{mujeeb2021}. This leads to poor performance across nearly all resamplers across datasets, as evidenced in both radar plots. On this classifier, our algorithm was unable to achieve the best performance, primarily because the classifier relies on random undersampling techniques that may discard important minority class data points. This approach contradicts the objective of our algorithm, which aims to generate additional data points after displacing noisy ones. 

We present the scatter plots and confusion matrices for the \textit{yeast} dataset—one of the sparsest and most imbalanced datasets with ten classes—in Fig.~\ref{fig: test_scatter2} and Fig.~\ref{fig: test_cm}, respectively. The t-SNE plots reveal that similar to the patterns observed with the NDE algorithm in Fig.~\ref{fig: test_scatter}, our NDESO algorithm generates data points with less overlap than other resamplers. In contrast, ECDNN produces relatively clean and separated data points but still exhibits significant overlaps. Additionally, the total number of classes, initially ten, is reduced to nine, indicating that ECDNN may have inadvertently removed critical data points. Furthermore, the confusion matrix plots highlight that our algorithm achieved the best performance, as evidenced by a clearer dark blue diagonal pattern, reflecting a strong alignment between predicted and actual values. This further confirms the accuracy and reliability of NDESO in correctly classifying the data. ADASYN, ENN, and KMeans-SMOTE were unsuccessful in resampling this dataset, which can also be observed in Table~\ref{tbl: mlp-results}, encountering the same error shown in Table~\ref{tbl: err-data}.

While our method shows a promising result, there is one notable caveat. Our algorithm calculates the average distance between a data point and its $k$-neighbors, then moves the data point closer to the centroid while maintaining the same distance. This distance calculation process can be computationally expensive, particularly for large datasets. As a result, when referring to Table~\ref{tbl: time-results}, our method does not show the best performance in terms of execution time. The time recorded reflects only the duration of the resampling process applied to the initial data before the classification task and does not represent the total execution time. More straightforward methods such as RandomUnder, RandomOver, NearMiss, and SMOTE, along with some of its variants, perform better in terms of time efficiency. These methods rely on more straightforward sampling techniques, especially RandomUnder and NearMiss, which can substantially reduce the majority class to balance the dataset. However, this comes at a cost—these methods produce fewer representative samples, causing lower test performance. On the other hand, while our method is not the fastest, it is not the slowest either. Specifically, our method outperforms the SMOTE-CDNN, a more recent SMOTE variant with relatively good resampling results. Additionally, our method is faster than several other methods, such as ECDNN, SVM-SMOTE, and K-means-SMOTE.
\section{Discussion}\label{sec: discussion}
The SMOTE method and its variations have been widely exploited and proven effective in addressing imbalanced class problems in most scenarios. However, while these methods generate quantitatively balanced data, the synthetic data patterns often deviate from the original distribution, introducing noise into the resampled dataset, especially in cases where there are sparse data with overlapping data points. Recently, \citep{Wang2023d} proposed a hybrid method called SMOTE-CDNN that combines undersampling and oversampling techniques. The undersampling technique in this method removes data points that do not match the class prediction upon the centroid displacement. However, this process may unintentionally discard important information, causing the oversampling to replicate more from less essential ones, potentially increasing the amount of non-representative data points.
In this study, we propose an alternative approach that preserves these noisy data by shifting them closer to the center of their class before performing random oversampling to balance the dataset. There are several benefits to this approach:
\begin{enumerate}
	\item \textbf{Relative positioning}\\
	Moving these noisy data points keeps their relative positioning within the cluster intact, preserving their contribution toward its class data distribution.
	\item \textbf{Noise reduction}\\
	Noisy data points are outliers that can distort the representation of their class. Moving them closer to their centroid helps align them with the central of their class, reducing their impact as noise.
	\item \textbf{Class boundaries}\\
	Moving them from overlapping data points increases the distinct boundaries between classes.
	\item \textbf{Statistical properties}\\
	Moving these noisy data points towards the center, the dataset's statistical properties (e.g., mean, variance) are largely preserved.
	\item \textbf{Oversampling}\\
	Doing oversampling after such adjustments makes the generated synthetic data points more likely to align with the class’s data characteristics.
\end{enumerate}
We have validated this hypothesis through various tests, which show that our method consistently outperforms most other resampling methods. Our method successfully handles datasets with an extensive variation in imbalance levels, ranging from ratios of 1:1, 1:94, and 1:164 to a very extreme ratio of 1:853. This ratio far exceeds the level tested in \citep{Wang2023d}, which reached a maximum ratio of 1:130. In addition, the data distribution results of our method show a much better ability to overcome overlap or noise problems compared to SMOTE-CDNN and other SMOTE variations.

Our experimental results show a promising approach to address the challenges of sparse and imbalanced text data. However, future research can delve deeper to evaluate its effectiveness and efficiencies in the context of big data. Big data processing often requires more complex and resource-intensive workflows, especially when dealing with distributed systems and network constraints, as highlighted in various studies on resampling techniques for large-scale datasets \citep{Bagui2021, Singh2023}. An interesting direction for further exploration is the application of resampling methods to binary big data formats, such as images, spatial datasets, or graph-based representations. This could open up new opportunities in geospatial analysis, network science, and high-dimensional data processing, paving the way for broader adoption and adaptation of the proposed approach in various real-world scenarios.

\section{Conclusion}\label{sec: conclusion}
This paper proposes a hybrid resampling method consisting of a noisy data point displacement approach with a random oversampling technique to handle imbalanced multiclass data. Our foundational approach performs noisy data point displacement by taking the average distance of a data point among its $k$-neighbors and repositioning it closer to the class centroid with equal distance. This procedure is repeated across overlapping data points, resulting in a cleaner class separation ready for oversampling. The method continues to refine the dataset by performing random oversampling to balance the data distribution. The approach was validated by confronting it with 14 other baseline resamplers on nine classifiers across 20 real-world datasets. Various parameter variations of the method have also been evaluated to demonstrate its robustness. Extensive testing results with statistical test confirmation have shown that our approach outperforms most baselines, highlighting our method's suitability for various real-world imbalanced classification tasks. Further research on the effectiveness of our strategy towards resampling in big data environments with other types of datasets is an open topic for investigation.

\subsection*{Data availability}
The data that support the findings of this study are openly available in GitHub at \url{https://github.com/goshlive/imbalanced-ndeso}.



\end{document}